\documentclass[runningheads]{llncs}
\usepackage{graphicx}

\usepackage{tikz}
\usepackage{comment} 
\usepackage{amsmath,amssymb} 
\usepackage{color}
\usepackage{mathrsfs}
\usepackage{algorithm}
\usepackage{algpseudocode}
\usepackage{multirow}
\usepackage{caption}
\usepackage{subcaption}
\newcommand\blfootnote[1]{%
	\begingroup
	\renewcommand\thefootnote{}\footnote{#1}%
	\addtocounter{footnote}{-1}%
	\endgroup
}
\newcommand{\etal}{\textit{et al}. }
\newcommand{\ie}{\textit{i}.\textit{e}. }

\begin{document}
	\pagestyle{headings}
	\mainmatter
	\def\ECCVSubNumber{768}  
	
	\title{Whole-Body Human Pose Estimation \\in the Wild}

	\titlerunning{Whole-Body Human Pose Estimation in the Wild}
	%
	\author{Sheng Jin\inst{1,2}\orcidID{0000-0001-5736-7434} \and
		Lumin Xu\inst{3,2} \and
		Jin Xu\inst{2} \and
		Can Wang\inst{2} \and \\
		Wentao Liu\inst{2\dagger} \and
		Chen Qian\inst{2} \and Wanli Ouyang\inst{4} \and Ping Luo\inst{1} \\[.21cm]
		$^{1}$ The University of Hong Kong \quad
		$^{2}$ SenseTime Research \\
		$^{3}$ The Chinese University of Hong Kong \quad
		$^{4}$ The University of Sydney \\
		\tt\small \{jinsheng, xulumin, wangcan, liuwentao, qianchen\}@sensetime.com \quad
		wanli.ouyang@sydney.edu.au, pluo@cs.hku.hk  }
	\authorrunning{S. Jin et al.}
	%
	\institute{}
	
	\maketitle
	
	\blfootnote{$^{\dagger}$Corresponding author.}
	
	\begin{abstract}
		This paper investigates the task of 2D human whole-body pose estimation, which aims to localize dense landmarks on the entire human body including face, hands, body, and feet. As existing datasets do not have whole-body annotations, previous methods have to assemble different deep models trained independently on different datasets of the human face, hand, and body, struggling with dataset biases and large model complexity. To fill in this blank, we introduce COCO-WholeBody which extends COCO dataset with whole-body annotations. To our best knowledge, it is the first benchmark that has manual annotations on the entire human body, including 133 dense landmarks with 68 on the face, 42 on hands and 23 on the body and feet. A single-network model, named ZoomNet, is devised  to take into account the hierarchical structure of the full human body to solve the scale variation of different body parts of the same person. ZoomNet is able to significantly outperform existing methods on the proposed COCO-WholeBody dataset. Extensive experiments show that COCO-WholeBody not only can be used to train deep models from scratch for whole-body pose estimation but also can serve as a powerful pre-training dataset for many different tasks such as facial landmark detection and hand keypoint estimation. The dataset is publicly available at \url{https://github.com/jin-s13/COCO-WholeBody}.
		\keywords{Whole-body human pose estimation, facial landmark detection, hand keypoint estimation}
	\end{abstract}
	
	\begin{figure}[t]
		\centering
		\includegraphics[width=0.85\textwidth]{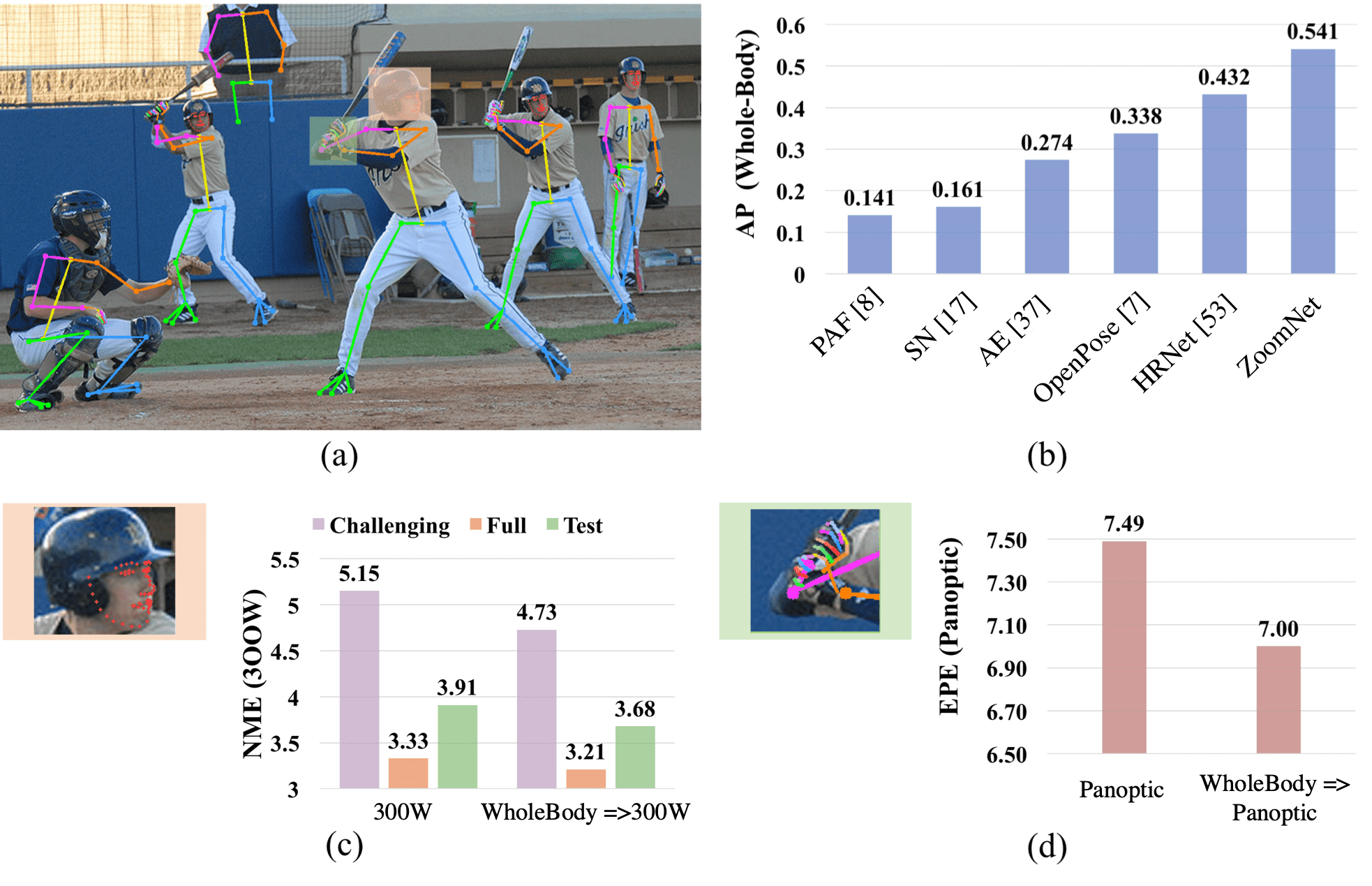}
		\caption{The proposed COCO-WholeBody dataset provides manual annotations of dense landmarks on the entire human body including body, face, hands, and feet. (a) visualizes an image as an example. The whole-body human pose estimation is challenging because different body parts have different variations such as scale. (b) shows that ZoomNet significantly outperforms the prior arts on this challenging task.
			(c) and (d) show that existing facial/hand landmark estimation algorithms can be improved by pretraining on COCO-WholeBody.
		}
		\label{fig:motivation}
	\end{figure}

	\section{Introduction}
	Human pose estimation has significant progress in the past few years. Recently, a more challenging task called \emph{whole-body} pose estimation is proposed and attracts much attention. As shown in Fig.~\ref{fig:motivation}a., whole-body pose estimation aims at localizing keypoints of body, face, hand, and foot simultaneously. This task is important for the development of downstream applications, such as virtual reality, augmented reality, human mesh recovery, and action recognition. 
	
	In recent years, deep neural networks (DNNs) become popular for keypoint estimation. However, to our knowledge, existing datasets of human pose estimation do not have manual annotations of the entire human body.
	Therefore, previous works trained their models separately on different datasets of face, hand and human body. For example, OpenPose~\cite{cao2017realtime} combines multiple DNNs trained independently on different datasets, including one DNN for body pose estimation on COCO~\cite{lin2014microsoft}, one DNN for face keypoint detection by combining many datasets (\emph{i.e.}  Multi-PIE~\cite{Gross2010Image}, FRGC~\cite{phillips2005overview} and i-bug~\cite{sagonas2013300}), and another DNN for hand keypoint detection on Panoptic~\cite{simon2017hand}. These methods may have several drawbacks. First, the data size of the current in-the-wild datasets of 2D hand keypoints is limited. Most approaches of hand pose estimation  have to use lab-recorded~\cite{tompson2014real,gomez2017large} or synthetic datasets~\cite{mueller2018ganerated,mueller2017real,sharp2015accurate}, hampering the performance of the existing methods in real-world scenarios. Second, the variations such as illumination, pose and scales in the existing human face~\cite{belhumeur2013localizing,koestinger2011annotated,le2012interactive,messer1999xm2vtsdb,sagonas2013300,zhu2012face}, hand~\cite{tompson2014real,gomez2017large,gomez2017large,mueller2018ganerated}, and body datasets~\cite{andriluka14cvpr,andriluka2018posetrack,wu2017ai,lin2014microsoft} are different, inevitably introducing dataset biases to the learned deep networks, thus hindering the development of algorithms to comprehensively consider the task as a whole. 
	
	To address the above issues, we propose a novel large-scale dataset for whole-body pose estimation, named COCO-WholeBody, which fully annotates the bounding boxes of face and hand, as well as the keypoints of face, hand, and foot for the images from COCO~\cite{lin2014microsoft}. To our knowledge, this is the first dataset that has whole-body annotations. COCO-WholeBody enables us to take into account the hierarchical structure of the human body and the correlations between different body parts to estimate the entire body pose. Therefore, it enables the development of a more reliable human body pose estimator. In addition, it will also stimulate productive research on related areas such as face and hand detection, face alignment and 2D hand pose estimation. The effectiveness of COCO-WholeBody is validated by using cross-dataset evaluation, which demonstrates that COCO-WholeBody can be used as a powerful pre-training dataset for various tasks, such as facial landmark localization and hand keypoint estimation. We overview the cross-dataset evaluations as shown in Fig.\ref{fig:motivation}c., d.
	
	The task of whole-body pose estimation has not been fully exploited in the literature because of missing a representative benchmark. Previous works \cite{cao2018openpose,hidalgo2019single} are mainly the bottom-up approaches, which simultaneously detect the keypoints for all persons in an image at once. They are generally efficient, however, they might suffer from scale variance of persons, causing inferior performance for small persons. Recent works \cite{sun2019deep,xiao2018simple} found that the top-down alternatives would have higher accuracy, because top-down methods normalize the human instances to roughly the same scale and are less sensitive to the scale variance of different human instances. However, to our knowledge, there is no existing top-down approach for whole-body pose estimation. With COCO-WholeBody, we are able to fill in this blank by designing a top-down whole-body pose estimator. However, predicting all the keypoints for whole-body pose estimation will lead to inferior performance, because the scales of human body, face and hand are different. For example, human body pose estimation requires a large receptive field to handle occlusion and complex poses, while face and hand keypoint estimation requires higher image resolution for accurate localization. If all the keypoints are treated equally and directly predicted at once, the performance is suboptimal.
	
	To solve this technical problem, we propose ZoomNet to effectively handle the scale variance in whole-body pose estimation. ZoomNet follows the top-down paradigm. Given a human bounding box of each person, ZoomNet first localizes the easy-to-detect body keypoints and estimates the rough position of hands and face. Then it zooms in to focus on the hand/face areas and predicts keypoints using features with higher resolution for accurate localization. Unlike previous approaches~\cite{cao2018openpose} which usually assemble multiple networks, ZoomNet has a single network that is end-to-end trainable. It unifies five network heads including the human body pose estimator, hand and face detectors, and hand and face pose estimators into a single network with shared low-level features. Extensive experiments show that ZoomNet outperforms the state-of-the-arts~\cite{cao2018openpose,hidalgo2019single} by a large margin, \emph{i.e.} $0.541$ vs $0.338$~\cite{cao2018openpose} for whole-body mAP on COCO-WholeBody.
	
	Our major contributions can be summarized as follows. \textbf{(1)} We propose the first benchmark dataset for whole-body human pose estimation, termed COCO-WholeBody, which encourages more exploration of this task. To evaluate the effectiveness of COCO-WholeBody, we extensively examine the performance of several representative approaches on this dataset. Also, the generalization ability of COCO-WholeBody is validated by cross-dataset evaluations, showing that COCO-WholeBody can serve as a powerful pre-training dataset for many tasks, such as facial landmark localization and hand keypoint estimation. 
	\textbf{(2)} We propose a top-down single-network model, ZoomNet to solve the scale variance of different body parts in a single person. Extensive experiments show that the proposed method significantly outperforms previous state-of-the-arts.
	
	\section{Related Work}
	\label{sec:related_work}
	
	\subsection{2D Keypoint Localization Dataset}
	As shown in Table~\ref{tab:dataset}, there are many datasets separately annotated for localizing the keypoints of body  \cite{andriluka2018posetrack,andriluka14cvpr,eichner2010we,lin2014microsoft,wu2017ai}, hand \cite{gomez2017large,mueller2018ganerated,simon2017hand,tompson2014real,yuan2017bighand2} or face \cite{belhumeur2013localizing,koestinger2011annotated,le2012interactive,messer1999xm2vtsdb,sagonas2013300,zhu2012face}. These datasets are briefly discussed in this section.
	
	\begin{table}[tb]
		\scriptsize
		\caption{Overview of some popular public datasets for 2D keypoint estimation in RGB images. Kpt stands for keypoints, and \#Kpt means the annotated number. ``Wild'' denotes whether the dataset is collected in-the-wild. * means head box.}
		\begin{center}
			\begin{tabular}{c|c|c|c|ccc|ccc|c}
				\hline
				DataSet & Images & \#Kpt & Wild &  Body   &    Hand    &    Face   &   Body   &    Hand    &   Face & Total \\ 
				&        &      &        & Box     &    Box     &    Box    &   Kpt    &    
				Kpt     &    Kpt & Instances \\\hline\hline
				\emph{MPII}~\cite{andriluka14cvpr}      & 25K & 16 & \checkmark &  \checkmark &   & * & \checkmark &    & & 40K \\
				\emph{MPII-TRB}~\cite{duan2019trb}      & 25K & 40 & \checkmark &  \checkmark &   & * & \checkmark &    & & 40K \\
				\emph{CrowdPose}~\cite{li2019crowdpose} & 20K & 14 & \checkmark  & \checkmark &   & & \checkmark & &   & 80K  \\ 
				\emph{PoseTrack}~\cite{andriluka2018posetrack} & 23K & 15 & \checkmark  & \checkmark &   & & \checkmark & &   & 150K  \\ 
				\emph{AI Challenger}~\cite{wu2017ai}            &  300K & 14 & \checkmark & \checkmark  &   &  & \checkmark &    &  & 700K  \\
				\emph{COCO}~\cite{lin2014microsoft}    & 200K & 17  & \checkmark & \checkmark  &   & * & \checkmark &    &  & 250K  \\\hline
				\emph{OneHand10K}~\cite{Yangang2018Mask}            & 10K & 21 &\checkmark  &    & \checkmark & &  & \checkmark & & - \\
				\emph{SynthHand}~\cite{mueller2017real}   & 63K &  21 &  &    & \checkmark & &  & \checkmark & & - \\
				\emph{RHD}~\cite{zb2017hand}            & 41K &  21 &  &    & \checkmark & &  & \checkmark & & - \\
				\emph{FreiHand}~\cite{Freihand2019}     & 130K & 21 &  &    &  & &  & \checkmark & & - \\
				\emph{MHP}~\cite{gomez2017large}        & 80K & 21  &  &    & \checkmark & &  & \checkmark & & - \\
				\emph{GANerated}~\cite{mueller2018ganerated}  & 330K & 21 &   &   &    & &  & \checkmark & & - \\
				\emph{Panoptic}~\cite{simon2017hand}    & 15K & 21 &  &   &  \checkmark & &  & \checkmark & & - \\ \hline
				\emph{WFLW}~\cite{wu2018look}    & 10K  &  98 & \checkmark  &   &   & \checkmark & &    & \checkmark & - \\
				\emph{AFLW}~\cite{koestinger2011annotated}    & 25K  & 19 & \checkmark  &   &   & \checkmark & &    & \checkmark & - \\
				\emph{COFW}~\cite{Burgos2013Robust}    & 1852 & 29 & \checkmark  &   &   & \checkmark & &    & \checkmark & - \\
				\emph{300W}~\cite{sagonas2013300}            & 3837 & 68 & \checkmark &   &   & \checkmark & & & \checkmark & -      \\ \hline
				COCO-WholeBody & 200K & 133 & \checkmark  & \checkmark & \checkmark   &\checkmark  &  \checkmark & \checkmark & \checkmark & 250K  \\ \hline	
			\end{tabular}
		\end{center}
		\label{tab:dataset}
	\end{table}
	
	\textbf{Body Pose Dataset.}
	There have been several body pose datasets~\cite{andriluka2018posetrack,andriluka14cvpr,duan2019trb,li2019crowdpose,lin2014microsoft,wu2017ai}. \emph{COCO} \cite{lin2014microsoft} is one of the most popular, which offers 17-keypoint annotations in uncontrolled conditions. Our COCO-WholeBody is an extension of COCO, with densely annotated 133 face/hand/foot keypoints. The task of whole-body pose estimation is more challenging, due to 1) higher localization accuracy required for face/hands and 2) scale variance between body and face/hands.
	
	\textbf{Hand Keypoint Dataset.} Most existing 2D RGB-based hand keypoint datasets are either synthetic~\cite{mueller2018ganerated,zb2017hand} or captured in the lab environment~\cite{gomez2017large,simon2017hand,Freihand2019}. For example, \emph{Panoptic}~\cite{simon2017hand} is a well-known hand pose estimation dataset, recorded in the CMU's Panoptic studio with multiview dome settings. However, it is limited to a controlled laboratory environment with a simple background. OneHand10K~\cite{Yangang2018Mask} is a recent in-the-wild 2d hand pose dataset. However, the size is still limited. Our COCO-WholeBody is complementary to these RGB-based hand keypoint datasets. It contains about 100K 21-keypoint labeled hands and hand boxes that are captured in unconstrained environment. To the best of our knowledge, it is the largest in-the-wild dataset for 2D RGB-based hand keypoint estimation. It is very challenging, due to occlusion, hand-hand interaction, hand-object interaction, motion blur, and small scales.
	
	\textbf{Face Keypoint Dataset.} Face keypoint datasets~\cite{Burgos2013Robust,koestinger2011annotated,sagonas2013300,wu2018look} play a crucial role for the development of facial landmark detection a.k.a. face alignment. Among them, \emph{300W}~\cite{sagonas2013300} is the most popular. It is a combination of LFPW~\cite{belhumeur2013localizing}, AFW~\cite{zhu2012face}, HELEN~\cite{le2012interactive}, XM2VTS~\cite{messer1999xm2vtsdb} with 68 landmarks annotated for each face image. Our proposed COCO-WholeBody follows the same annotation settings as 300W and 68 keypoints for each face are annotated. Compared to 300W, COCO-WholeBody is much larger and is more challenging as it contains more blurry and small-scale facial images (see Fig~\ref{fig:face_hand_compare}a.). 
	
	\textbf{DensePose Dataset.}
	Our work is also related to DensePose~\cite{alp2018densepose} which provides a dense 3D surface-based representation for human shape. However, since the keypoints in DensePose are uniformly sampled, they lack specific joint articulation information and details of face/hands are missing.
	
	\subsection{Keypoints Localization Method}
	\textbf{Body Pose Estimation.} Recent multi-person body pose estimation approaches can be divided into bottom-up and top-down approaches. Bottom-up approaches \cite{cao2017realtime,Insafutdinov2016ArtTrack,Insafutdinov2016DeeperCut,Iqbal2016PoseTrack,jin2019multi,jin2017towards,newell2017associative,nie2017generative,papandreou2018personlab,pishchulin2016deepcut} first detect all the keypoints of every person in images and then group them into individuals. Top-down methods~\cite{chen2018cascaded,fang2017rmpe,he2017mask,liu2018cascaded,newell2016stacked,papandreou2017towards,sun2019deep,xiao2018simple} first detect the bounding boxes and then predict the human body keypoints in each box. By resizing and cropping, top-down approaches normalize the poses to approximately the same scale. Therefore, they are more robust to human-level scale variance and recent state-of-the-arts are obtained by top-down approaches. However, direct usage of the top-down methods for whole-body pose estimation will encounter the problem of scale variance of different body parts (body vs face/hand). To tackle this problem, we propose ZoomNet, a single-network top-down approach that zooms in to the hand/face regions and predicts the hand/face keypoints using higher image resolution for accurate localization.
	
	\textbf{Face/Hand/Foot Keypoint Localization.} 
	Previous works mostly treat the tasks of face/hand/foot keypoint localization independently and solve by different models. For facial keypoint localization, cascaded networks~\cite{cao2014face,sun2013deep,tzimiropoulos2015project,xiong2013supervised} and multi-task learning~\cite{trigeorgis2016mnemonic,zhang2015learning} are widely adopted. For hand keypoint estimation, most work rely on auxiliary information such as depth information~\cite{oikonomidis2012tracking,sharp2015accurate,sridhar2015fast} or multi-view~\cite{guan2006multi,neverova2014multi} information. For foot keypoint estimation, Cao \emph{et al.}~\cite{cao2018openpose} proposed a generic bottom-up method. In this paper, we propose ZoomNet to solve the tasks of face/hand/foot keypoint localization as a whole. It takes into account the inherent hierarchical structure of the full human body to solve the scale variation of different parts in the same person.
	
	\textbf{Whole-Body Pose Estimation.} Whole-body pose estimation has not been well studied in the literature due to the lack of a representative benchmark. OpenPose~\cite{cao2018openpose,cao2017realtime,simon2017hand} applies multiple models (body keypoint estimator) to handle different kinds of keypoints. It first detects body and foot keypoints, and estimates the hand and face position. Then it applies extra models for face and hand pose estimation. Since OpenPose relies on multiple networks, it is hard to train and suffers from increased runtime and computational complexity. Unlike OpenPose, our proposed ZoomNet is a “single-network” method as it integrates five previously separated models (human body pose estimator, hand/face detectors, and hand/face pose estimators) into a single network with shared low-level features. Recently, Hidalgo \etal proposes an elegant method SN~\cite{hidalgo2019single} for bottom-up whole-body keypoint estimation. SN is based on PAF~\cite{cao2017realtime} which predicts the keypoint heatmaps for detection and part affinity maps for grouping. Since there exists no such dataset with whole-body annotations, they used a set of different datasets and carefully designed the sampling rules to train the model. However, bottom-up approaches cannot handle scale variation problem well and would have difficulty in detecting face and hand keypoints accurately. In comparison, our ZoomNet is a top-down approach that well handles the extreme scale variance problem. Recent works~\cite{joo2018total,romero2017embodied,xiang2019monocular} also explore the task of monocular 3D whole-body capture. Romero \etal proposes a generative 3D model~\cite{romero2017embodied} to express body and hands. Xiang \etal introduces a 3D deformable human model~\cite{xiang2019monocular} to reconstruct whole-body pose and Joo \etal presents Adam~\cite{joo2018total} which encompasses the expressive power for body, hands, and facial expression. Their methods still rely on OpenPose~\cite{cao2018openpose} to localize 2d body keypoints in images. 
	
	\section{COCO-WholeBody Dataset}
	COCO-WholeBody is the first large-scale dataset with the whole-body pose annotation available, to the best of our knowledge. In this section, we will describe the annotation protocols and some informative statistics.

	\begin{figure*}[tb]
		\centering
		\includegraphics[width=0.95\textwidth]{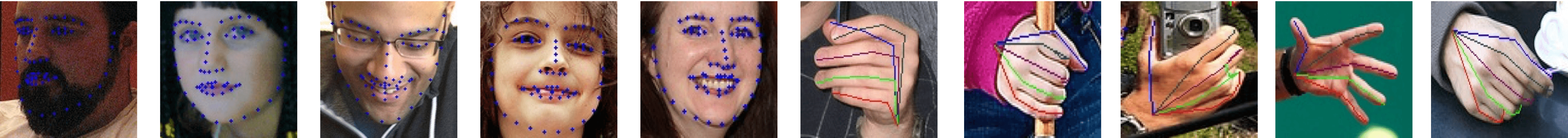}
		\caption{Annotation examples for face/hand keypoints in COCO-WholeBody.}
		\label{fig:anno_hand_face_example}
	\end{figure*}

	\begin{figure}[tb]
		\centering
		\begin{subfigure}[b]{0.4\textwidth}
			\centering
			\includegraphics[width=0.85\textwidth]{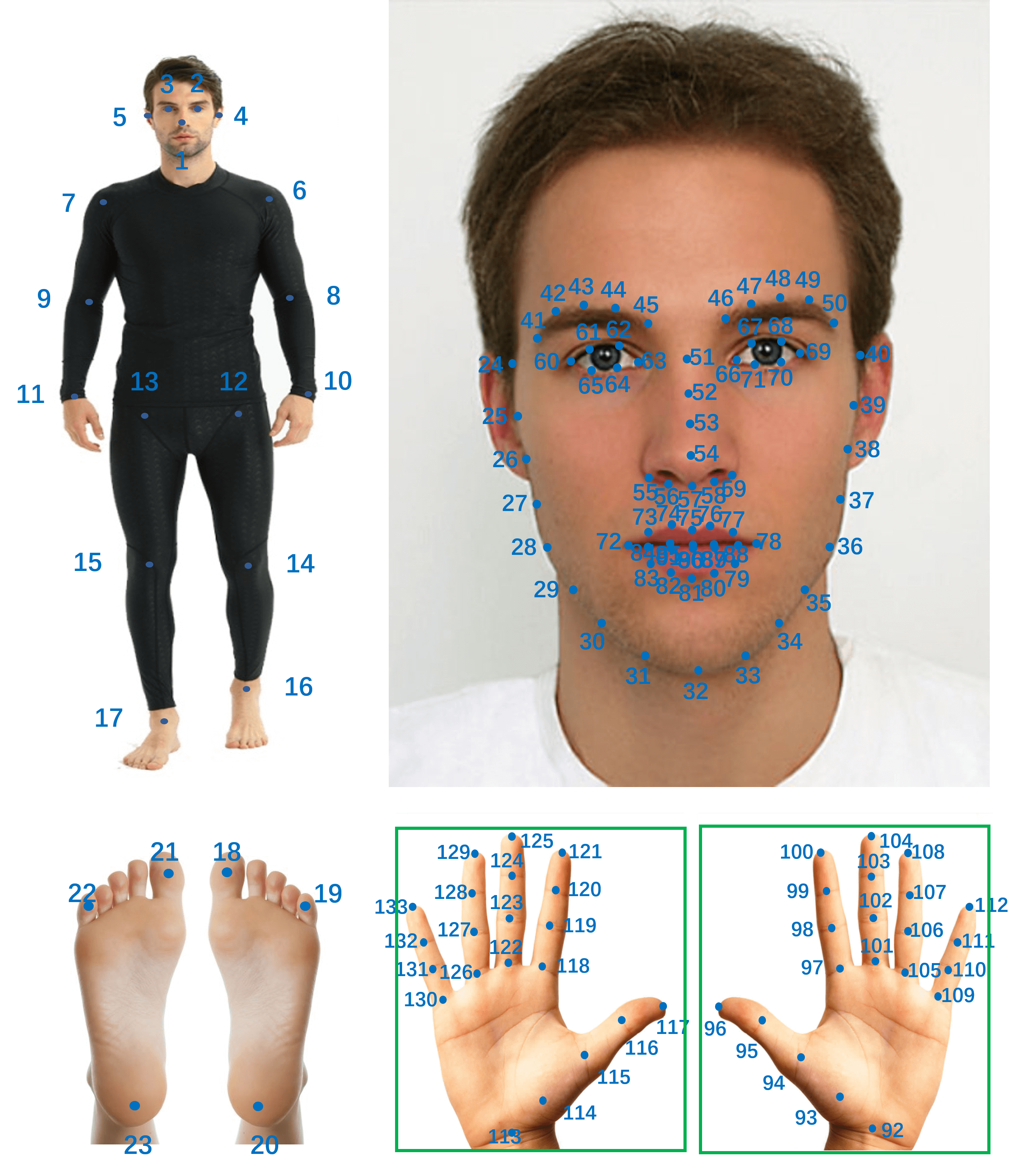}
			\caption{}
			\label{fig:anno_kpt}
		\end{subfigure}
		\begin{subfigure}[b]{0.55\textwidth}
			\centering
			\includegraphics[width=0.99\textwidth]{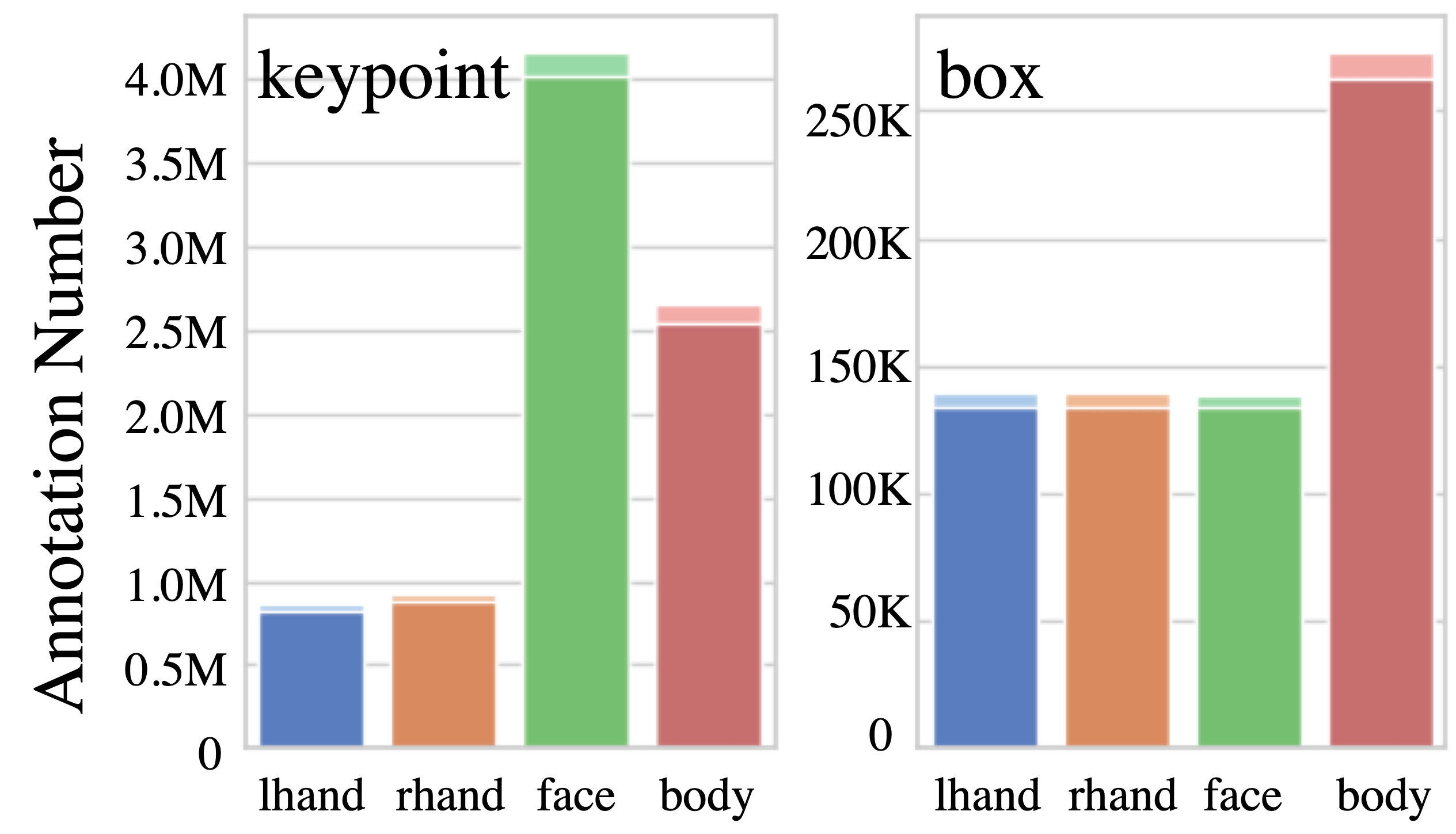}
			\caption{}
			\label{fig:number_statistics}
		\end{subfigure}
		\caption{(a) COCO-WholeBody annotation for 133 keypoints. (b)Statistics of COCO-WholeBody. The number of annotated keypoints and boxes of left hand (lhand), right hand (rhand), face and body are reported.}
	\end{figure}

	\subsection{Data Annotation}
	
	We annotate the face, hand and foot keypoints on the whole train/val set of COCO~\cite{lin2014microsoft} dataset and form the whole-body annotations with the original body keypoint labels together (see Fig.~\ref{fig:anno_hand_face_example}). For each person, we annotate 4 types of bounding boxes (person box, face box, left-hand box, and right-hand box) and 133 keypoints (17 for body, 6 for feet, 68 for face and 42 for hands). The face/hand box is defined as the minimal bounding rectangle of the keypoints. The keypoint annotations are illustrated in Fig.~\ref{fig:anno_kpt}. The face/hand boxes are labeled as \emph{valid}, only if the face/hand images are clear enough for keypoint labeling. Invalid boxes may be blurry or severely occluded. We only label keypoints for \emph{valid} boxes. Manual annotation for whole-body poses in an uncontrolled environment, especially for massive and dense hand and face keypoints, requires trained experts and enormous workload. As a rough estimate, the manual labeling cost of a professional annotator is up to: 10 min/face, 1.5 min/hand, and 10 sec/box. To speed up the annotation process, we follow the semi-automatic methodology to use a set of pre-trained models (for face and hand separately) to pre-annotate and then conduct manual correction. Foot keypoints are directly manually labeled, since its labeling cost is relatively low. Specifically, the annotation process contains the following steps:
	\begin{enumerate}
		\item For each individual person, we manually label the face box, the left-hand box, and the right-hand box. The validity of the boxes is also labeled.
		\item Quality control. The annotation quality of the boxes is guaranteed through the strict quality inspection performed by another group of the annotators.
		\item For each valid face/hand box, we use pre-trained face/hand keypoint detectors to produce pseudo keypoint labels. We use a combination of the publicly available datasets to train a robust face keypoint detector and a hand keypoint detector based on HRNetV2~\cite{sun2019deep}.
		\item Manual correction of pseudo labels and further quality control. About 28\% of the hand keypoints and 6\% of the face keypoints are labeled as invalid and manually corrected by human annotators. By using the semi-automatic annotation, we saw about $89\%$ reduction in the time required for annotation. 
	\end{enumerate}
	To measure the annotation quality, we also had 3 annotators to label the same batch of 500 images for face/hand/foot keypoints. The standard deviation of the human annotation is calculated for each keypoint (see Fig.~\ref{fig:error_analyze}a.), which is used to calculate the normalized factor of whole-body keypoint for evaluation. For ``body keypoints'', we directly use the standard deviation reported in COCO~\cite{lin2014microsoft}.

	\subsection{Evaluation Protocol and Evaluation Metrics}
\begin{figure*}[tb]
		\centering
		\includegraphics[width=0.85\linewidth]{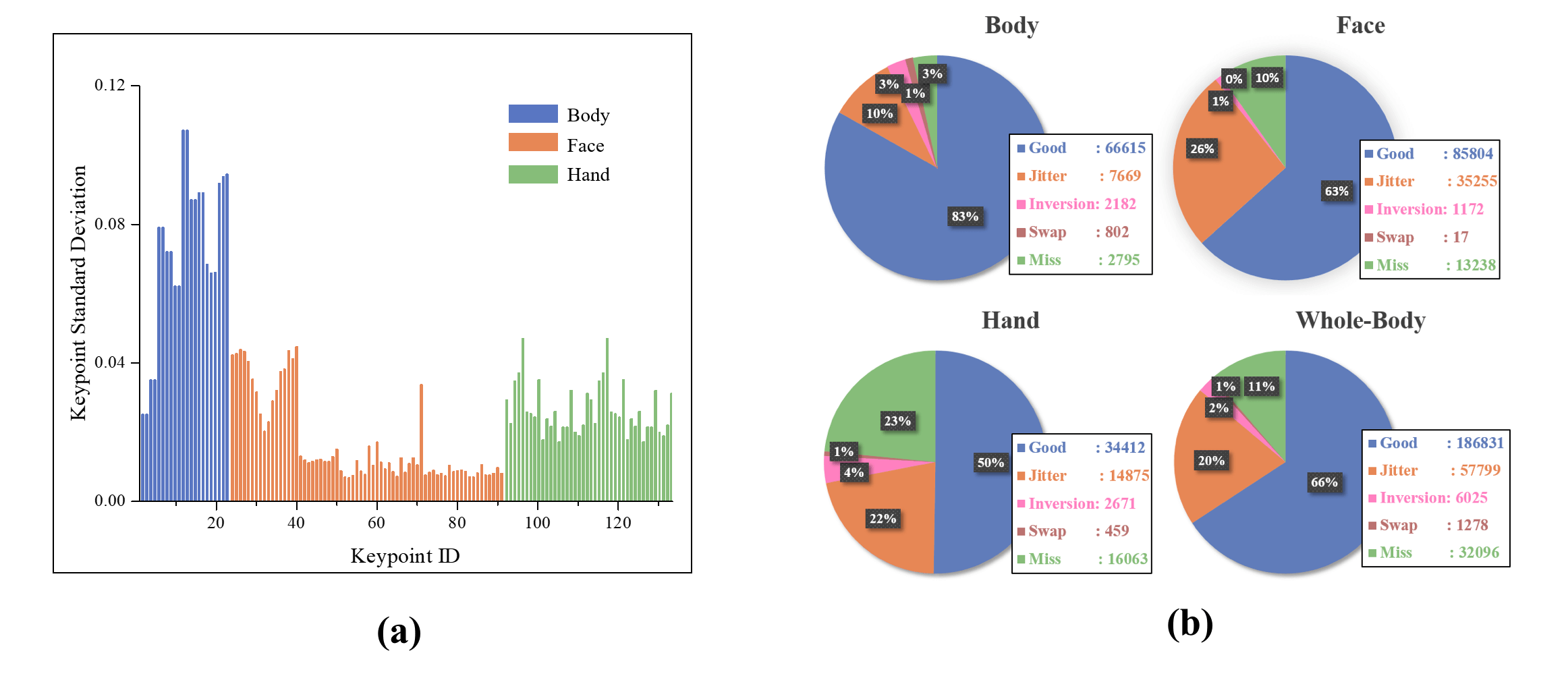}
		\caption{(a) The normalized standard deviation of manual annotation for each keypoint. Body keypoints have larger manual annotation variance than face and hand keypoints. (b) An example of error diagnosis results of ZoomNet for whole-body pose estimation: jitter, inversion, swap and missing.}
		\label{fig:error_analyze}
	\end{figure*}

	The evaluation protocol of whole-body pose estimation follows the current practices in the literature~\cite{lin2014microsoft,wu2017ai}. All algorithms are trained on COCO-WholeBody training set and evaluated on COCO-WholeBody validation set. We use mean Average Precision (mAP) and Average Recall (mAR) for evaluation, where Object Keypoint Similarity (OKS) is used to measure the similarity between the prediction and the ground truth poses. Invalid boxes and keypoints are masked out during both training and evaluation, thus not affecting the results. The ignored regions are masked out, and only visible keypoints are considered during evaluation. As shown in Fig.~\ref{fig:error_analyze}b., we also develop a tool for deeper performance analysis based on~\cite{Ronchi2017Benchmarking} which will be provided to facilitate offline evaluation. 
	
	\begin{figure*}[tb]
		\centering
		\includegraphics[width=0.98\textwidth]{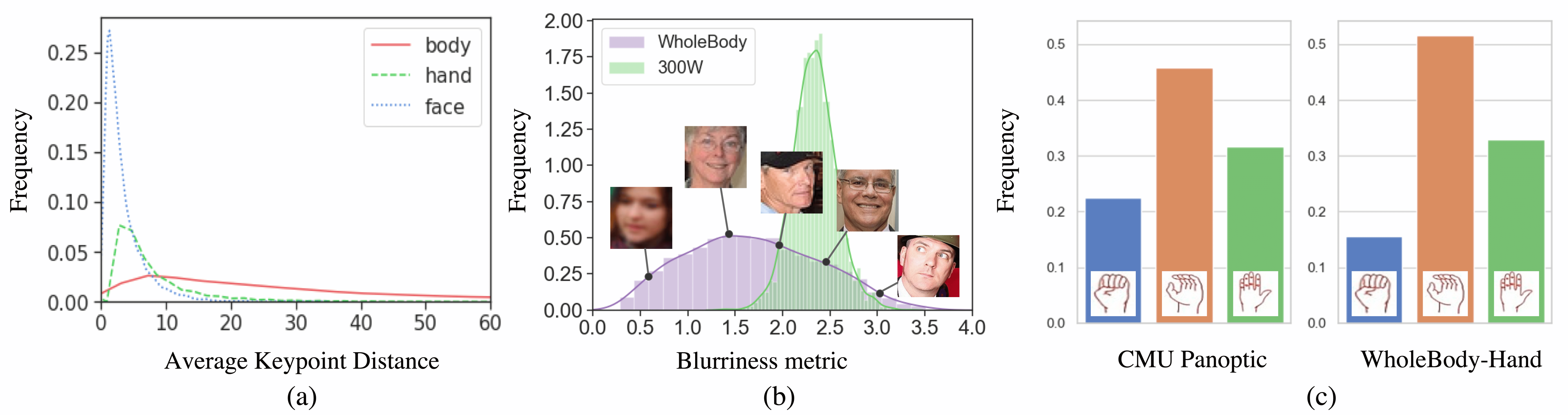}
		\caption{COCO-WholeBody is challenging as it contains (a) large ``scale variance'' of body/face/hand, measured by the average keypoint distance, (b) more blurry face images than 300W and (c) more complex hand poses than Panoptic.}
		\label{fig:face_hand_compare}
	\end{figure*}

	\subsection{Dataset Statistics}
	\textbf{Dataset Size.} 
	COCO-WholeBody is a large-scale dataset with keypoint and bounding box annotations. The number of annotated keypoints as well as boxes of left hand (lhand), right hand (rhand), face and body are shown in Fig.~\ref{fig:number_statistics}. About 130K face and left/right hand boxes are labeled, resulting in more than 800K hand keyponits and 4M face keypoints in total.
	
	\textbf{Scale Difference.}
	Distribution of the average keypoint distance of different parts in WholeBody Dataset is summarized in Fig.~\ref{fig:face_hand_compare}a. We calculate the distance between keypoint pairs in the tree-structured skeleton. Hand/face have obviously much smaller scales than body. The various scale distribution makes it challenging to localize keypoints of different human parts simultaneously.

	\textbf{Facial Image ``Blurriness''.}
	Face image ``blurriness'' is a key factor for facial landmark localization. We choose a variation of the Laplacian method~\cite{Pech2000Diatom} to measure it. Specifically, an image is first converted into a grayscale image and resized into $112 \times 112$. The log10 of the Laplacian of the converted image is used as the ``blurriness'' measurement (the higher the better).  The distribution of the blurriness is shown in Fig.~\ref{fig:face_hand_compare}b. We find that most facial images fall in the interval between $1$ and $3$ and are clear enough for accurate keypoint localization. Compared with 300W~\cite{sagonas2013300}, WholeBody has a larger variance of blurriness and contains more challenging images (blurriness $<1$).
	
	\textbf{Gesture Variances for Hands.}
	We first normalize the 2D hand poses by rotating and scaling and then cluster them into three main categories: ``fist'', ``palm'' and ``others''. Unlike most previous hand datasets that are collected in constrained environments, our WholeBody-Hand is collected in-the-wild. Compared with Panoptic~\cite{gomez2017large}, WholeBody-Hand is more challenging as it contains a larger proportion of hand images grasping or holding objects.
	
	Overall, COCO-WholeBody is a large-scale dataset with great diversity, which will not only promote researches on the whole-body pose estimation but also contribute to other related areas, such as face and hand keypoint estimation.
	
	\section{ZoomNet: Whole-Body Pose Estimation}
	
	\begin{figure*}[tb]
		\centering
		\includegraphics[width=0.98\textwidth]{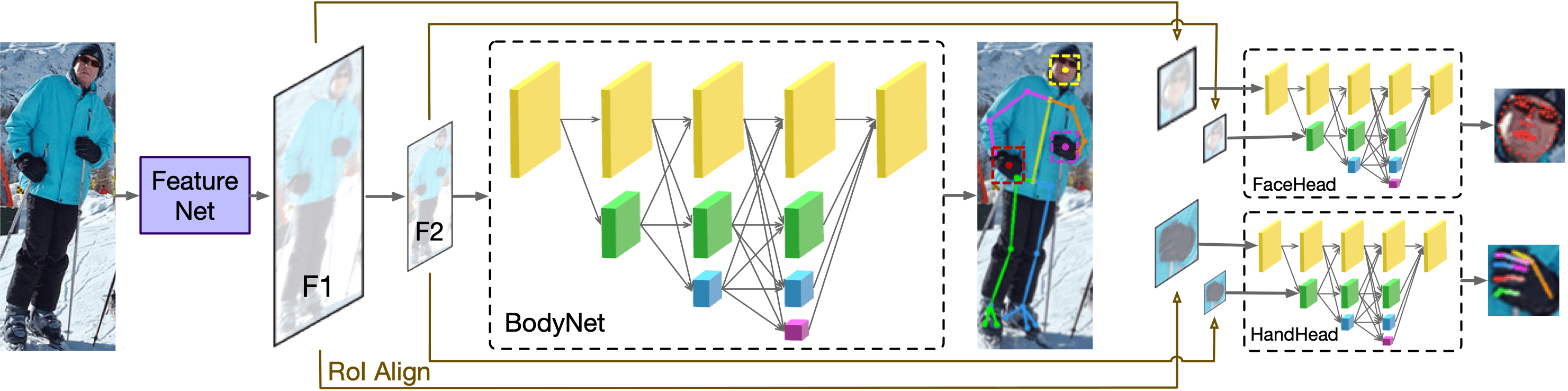}
		\caption{ ZoomNet is a single-network model, which consists of FeatureNet, BodyNet and Face/HandHead. FeatureNet extracts low-level shared features for BodyNet and Face/HandHead. BodyNet predicts body/foot keypoints and the approximate regions of face/hands, while Face/HandHead zooms in to these regions and predict face/hand keypoints with features of higher resolution.
		}
		\label{fig:architecture}
	\end{figure*}
	
	In this section, we will introduce our whole-body pose estimation pipeline. Given an RGB image, we follow~\cite{xiao2018simple,sun2019deep} to use an off-the-shelf FasterRCNN~\cite{renNIPS15fasterrcnn} human detector to generate human body candidates. For each human body candidate, ZoomNet localizes the whole-body keypoints. As shown in Fig.~\ref{fig:architecture}, ZoomNet predicts body/foot keypoints and face/hand keypoints successively in a single network, consisting of the following submodules:
	
	\textbf{FeatureNet}: the input image is processed by FeatureNet to extract shared features ($F1$ and $F2$). It consists of two convolutional layers, each of which downsamples the corresponding input to 1/2 resolution, and a bottleneck block for effective feature learning. The input image size is $384 \times 288$ and the output feature map sizes for $F1$ and $F2$ are $192 \times 144$ and $96 \times 72$, respectively.
	
	\textbf{BodyNet}: using the features extracted from FeatureNet, BodyNet predicts body/foot keypoints and face/hand bounding boxes at the same time. Each bounding box is represented by four corner points and one center point. In total, 38 keypoints are generated for each person simultaneously. BodyNet is a multi-resolution network with 38 output channels. 
	
	\textbf{HandHead and FaceHead}: Using face and hand bounding boxes predicted by BodyNet, we crop the features in the corresponding areas from F1 and F2. The features from F1 are resized to $64 \times 64$ and those from F2 are resized to $32 \times 32$. Then HandHead and FaceHead are applied to predict the heatmaps of face/hand keypoints with the output resolution of $64 \times 64$ in parallel. 
	
	ZoomNet can be based on any state-of-the-art network architecture. In our implementation, we choose HRNet-W32~\cite{sun2019deep} as the backbone of BodyNet and HRNetV2p-W18~\cite{sun2019high} as the backbone of FaceHead/HandHead. Please refer to Supplementary for more implementation details.
	
	\subsection{Localizing body keypoints and face/hand boxes with BodyNet}
	Our face/hand box localization is inspired by CornerNet~\cite{law2018cornernet}, which represents the object with keypoint pairs and designs a one-stage keypoint-based detector. In our case, each person has three types of bounding boxes to predict: the face box, the left-hand box, and the right-hand box. Four corner points and one center point are used to represent a box. We use 2D confidence heatmaps to encode both the human body keypoints and the corner keypoints. During inference, the bounding box is obtained by the closest bounding box of the 4 corner points.
	
	\subsection{Face/hand keypoint estimation with HandHead and FaceHead}
	Given the face/hand bounding boxes predicted by BodyNet, RoIAlign~\cite{he2017mask} is applied to extract the features of the face/hand areas from the feature maps $F1$ and $F2$ of FeatureNet. The corresponding visual features are cropped and up-scaled to a higher resolution. With the extracted features, HandHead and FaceHead are used for face and hand keypoint estimation. HandHead and FaceHead use the same network architecture (HRNet-W18). The features extracted by RoIAlign are processed by the HandHead and FaceHead separately. In this way, we are able to preserve the high-resolution for the hand/face regions, and larger receptive fields for body keypoint estimation at the same time. 
	
	\section{Experiments}
	
	\subsection{Evaluation on COCO-WholeBody Dataset}
	To the best of our knowledge, there are only two existing approaches that target at the 2D whole-body pose estimation task, \ie OpenPose~\cite{cao2018openpose} and SN~\cite{hidalgo2019single}. To extensively evaluate the performance of the existing methods on the proposed COCO-WholeBody Dataset, we also build upon the existing multi-person human body pose estimation approaches, including both bottom-up (\ie Partial Affinity Field (PAF)~\cite{cao2017realtime} and Associate Embedding (AE)~\cite{newell2017associative}) and top-down methods (\ie HRNet~\cite{sun2019deep}), and adapt them to the more challenging whole-body pose estimation task using official codes (see Supplementary for more details). For fair comparisons, we retrain all methods on COCO-WholeBody and evaluate their performance with single-scale testing as shown in Table~\ref{tab:compare}. We show that our proposed ZoomNet outperforms them by a large margin. 
	
	Among these methods, SN~\cite{hidalgo2019single}, PAF~\cite{cao2017realtime}, AE~\cite{newell2017associative} and HRNet~\cite{sun2019deep} follow a one-stage paradigm and predict all the keypoints simultaneously. Interestingly, we find that in the task of whole-body pose estimation, directly learning to predict all 133 keypoints simultaneously, including body, face, hand keypoints, may harm the original body keypoint estimation accuracy. In Table~\ref{tab:compare}, ``-body'' means that we only train the model on the original COCO-body keypoint  (17 keypoints). We compare the body keypoint estimation results of the model learning the whole-body keypoints versus the model learning the body keypoints only. We observe considerable accuracy decrease by comparing PAF vs PAF-body (-14.3\% mAP and -14.2\% mAR), AE vs AE-body(-17.7\% mAP and -17.0\% mAR) and HRNet vs HRNet-body(-9.9\% mAP and -10.0\% mAR). In comparison, our proposed ZoomNet uses a two-stage framework, which decouples the body keypoint estimation and face/hand keypoint estimation. The accuracy of body keypoint estimation is less affected (-1.5\% mAP and -0.7\% mAR).
	
	HRNet~\cite{sun2019deep} can be viewed as the \emph{one-stage} alternative of ZoomNet, since they share the same network backbone (HRNet-W32). ZoomNet significantly outperforms HRNet by 10.9\% mAP and 13.8\% mAR, demonstrating the effectiveness of the ``zoom-in'' design for solving the scale variation.
	
	OpenPose~\cite{cao2018openpose} is a \emph{multi-model} approach, where the hand/face model and the body model are not jointly trained, leading to sub-optimal results. In addition, the hand/face boxes of OpenPose are roughly estimated by hand-crafted rules from the estimated body keypoints. Therefore, the accuracy of the hand/face boxes is limited, which will hinder hand/face pose estimation. 
	
	\begin{table}[t]
		\scriptsize
		\caption{Whole-body pose estimation results on COCO-WholeBody dataset. For fair comparisons, results are obtained using single-scale testing.}
		\begin{center}
			\begin{tabular}{l|cc|cc|cc|cc|cc}
				\hline
				Method & \multicolumn{2}{c|}{body}  & \multicolumn{2}{c|}{foot}  & \multicolumn{2}{c|}{face}  & \multicolumn{2}{c|}{hand} & \multicolumn{2}{c}{whole-body} \\
				\cline{2-11}
				&  AP     & AR     & AP   & AR     &  AP  & AR     & AP    & AR   &  AP     & AR  \\
				\hline
				OpenPose~\cite{cao2018openpose} & 0.563 & 0.612 & 0.532 & 0.645 & 0.482 & 0.626 & 0.198 & 0.342 & 0.338 & 0.449 \\ 
				SN~\cite{hidalgo2019single} & 0.280 & 0.336 & 0.121 & 0.277 & 0.382 & 0.440 & 0.138 & 0.336 & 0.161 & 0.209 \\ 
				PAF~\cite{cao2017realtime} & 0.266 & 0.328 & 0.100 & 0.257 & 0.309 & 0.362 & 0.133 & 0.321 & 0.141 & 0.185 \\ 
				PAF-body~\cite{cao2017realtime} & 0.409 & 0.470 & - & - & - & - & - & - & - & - \\ 
				AE~\cite{newell2017associative} & 0.405 & 0.464 & 0.077 & 0.160 & 0.477 & 0.580 & 0.341 & 0.435 & 0.274 & 0.350 \\
				AE-body~\cite{newell2017associative} & 0.582 & 0.634 & - & - & - & - & - & - & - & - \\ \hline
				HRNet~\cite{sun2019deep}  & 0.659 & 0.709 & 0.314 & 0.424 & 0.523 & 0.582 & 0.300 & 0.363 & 0.432 & 0.520 \\
				HRNet-body~\cite{sun2019deep} & \textbf{0.758} & \textbf{0.809} & - & - & - & - & - & - & - & -  \\ \hline
				ZoomNet & 0.743 & 0.802 & \textbf{0.798} & \textbf{0.869} & \textbf{0.623} & \textbf{0.701} & \textbf{0.401} & \textbf{0.498} & \textbf{0.541} & \textbf{0.658}  \\
				\hline
			\end{tabular}
		\end{center}
		\label{tab:compare}
	\end{table}
	
	\textbf{Model complexity analysis.}
	The model complexity of ZoomNet is 27.36G Flops. By contrast, the model complexity of OpenPose~\cite{cao2018openpose} is 451.09G Flops in total (137.52G for BodyNet, 106.77G for FaceNet and $103.40\times 2=206.80$G for HandNet), and that of SN~\cite{hidalgo2019single} is 272.30G Flops. We also report the average runtime cost on COCO-WholeBody on one GTX-1080 GPU. SN is about 215.5ms/image, while ZoomNet is about 174.7ms/image on average (including a Faster RCNN human detector which takes about 106ms/image).
	
	\subsection{Cross-dataset Evaluation}
	In this section, we show that the proposed COCO-WholeBody is complementary to other separately labeled benchmarks by evaluating its generalization ability.
	
	\textbf{WholeBody-Face (WBF) Dataset.}
	We build WholeBody-Face (WBF) by extracting cropped face images/annotations from COCO-WholeBody. We conduct experiments on 300W~\cite{sagonas2013300} benchmark. We follow the common settings~\cite{sun2019high} to train models on 3,148 training images, validate on the ``common'' set and evaluate on the ``challenging'', ``full'' and ``test'' sets. We use the normalized mean error (NME) for evaluation and inter-ocular distance as normalization. The results are shown in Table~\ref{table:comparison_300w_fullset}. HR-Ours is our implementation of HRNetV2-W18~\cite{sun2019high} (HR). $^{*}$HR-Ours is obtained by training HR on WBF only and directly testing on 300W, which already outperforms RCN~\cite{honari2016recombinator}. After finetuning on 300W, it gets significantly better performance on ``challenging'' (4.73 vs 5.15), ``full'' (3.21 vs 3.33) and ``test'' (3.68 vs 3.91) than the prior arts.
	
	\renewcommand{\arraystretch}{1.3}
	\begin{table}[t]
		\caption{(a) Facial landmark localization (NME) on $300$W: ``common'' (for val), ``challenging'', ``full'' and ``test''. $^{*}$ means only training on WBF. $\downarrow$ means lower is better. (b) Cross-dataset evaluation results of HR. Different training and testing settings are evaluated on two datasets: WBH and Panoptic (Pano.)~\cite{simon2017hand}.}
		\begin{subtable}[b]{.5\textwidth}
			\centering
			\scriptsize
			\begin{tabular}{ l|ccccc  }
				\hline\noalign{\smallskip}
				& extra. &comm. $\downarrow$ & chall. $\downarrow$ & full $\downarrow$ & test $\downarrow$ \\ \hline
				\hline
				RCN \cite{honari2016recombinator} & - & $4.67$ & $8.44$ & $5.41$ &- \\
				DAN \cite{kowalski2017deep} & - & $3.19$ & $5.24$ & $3.59$ & $4.30$\\
				DCFE \cite{valle2018deeply} & w/$3$D & $\mathbf{2.76}$ & $5.22$ & $3.24$ & $3.88$ \\
				LAB  \cite{wu2018look} & w/B & $2.98$ & $5.19$ & $3.49$ & -\\
				HR~\cite{sun2019high} & -  &$2.87$ & $5.15$ & $3.32$ & $3.85$ \\
				\hline
				$^{*}$HR-Ours & -  & $4.61$ & $7.50$ & $5.17$ & $5.66$ \\
				HR-Ours & -  &$2.89$ & $5.15$ & $3.33$ & $3.91$ \\ 
				HR-Ours & WBF  &$2.84$ & $\mathbf{4.73}$ & $\mathbf{3.21}$ & $\mathbf{3.68}$ \\
				\hline
			\end{tabular}
			\caption{}
			\label{table:comparison_300w_fullset}
		\end{subtable}
		\quad
		\begin{subtable}[b]{.5\textwidth}
			\scriptsize
			\centering
			\begin{tabular}{c|cc|cc}
				\hline\noalign{\smallskip}
				\#&{Train-set} & {Test-set} & EPE $\downarrow$ & NME $\downarrow$ \\
				\noalign{\smallskip}
				\hline \hline
				\noalign{\smallskip}
				1&Pano. & Pano.  & ${7.49}$ & ${0.68}$ \\
				2&WBH $\Rightarrow$ Pano.  & Pano. & $\mathbf{7.00}$ & $\mathbf{0.63}$ \\ \hline
				3&WBH & WBH  & ${2.76}$ & ${6.66}$ \\
				4&Pano. $\Rightarrow$  WBH & WBH  & $\mathbf{2.70}$ & $\mathbf{6.49}$ \\ \hline
			\end{tabular}
			\caption{}
			\label{tab:panoptic}
		\end{subtable}
		\label{tab:face_and_hand_cross_dataset}
	\end{table}
	
	\textbf{WholeBody-Hand (WBH) Dataset.}
	For hand pose estimation, we experiment with HRNetV2-W18 (HR) on CMU Panoptic~\cite{simon2017hand} (Pano.), which is a standard benchmark for hand keypoint localization. We randomly split Pano~\cite{simon2017hand} by a rule of 70\%-30\% for training and validation. We report both the end-point-error (EPE) and the normalized mean error (NME) for evaluation. In NME, the hand bounding box is used as normalization. As shown in Table~\ref{tab:panoptic}, we analyze the generalization ability of WholeBody-Hand (WBH) by comparing the (1) HR trained on Pano., (2) HR pretrained on WBH and then finetuned on Pano., (3) HR trained on WBH, and (4) HR pretrained on Pano. and then finetuned on WBH. Comparing \#1 and \#2, we observe that pretraining on WBH brings about $6.5$\% improvement (from $7.49$ to $7.00$) in EPE on Pano. Comparing \#1 and \#3, we find that WBH vs Pano. is ($6.66$ vs $0.68$) NME and ($2.76$ vs $7.49$) EPE, when training/testing on its own dataset. This implies that the proposed WBH is much more challenging and that hand scales in WBH are smaller.
	
	\subsection{Analysis}
	\textbf{Effect of the bounding box accuracy on the keypoint estimation.} We experiment by replacing our predicted face/hand bounding boxes with the ground-truth bounding boxes and re-run our FaceHead/HandHead of ZoomNet to obtain the final face/hand keypoint detection result. As shown in table \ref{tab:personbbx}, using ground truth bounding boxes (Oracle) significantly improves the mAP of face/hand/whole-body by 19.6\%, 8.4\% and 23.6\% respectively. 
	
	\renewcommand{\arraystretch}{1.3}
	\begin{table}[t]
		\caption{Effect of bounding box accuracy on keypoint estimation, where Oracle means using gt boxes. (b) Effect of person scales on whole-body pose estimation.}
		\begin{subtable}[b]{.5\textwidth}
			\centering
			\scriptsize
			\begin{tabular}{l|cc|cc|cc}
				\hline
				Method & \multicolumn{2}{c|}{face}  & \multicolumn{2}{c|}{hand} & \multicolumn{2}{c}{whole-body} \\
				\cline{2-7}
				&    AP  & AR     & AP    & AR   &  AP     & AR  \\
				\hline
				Oracle  & 0.819 & 0.854 & 0.485 & 0.578 & 0.777 & 0.856   \\
				Ours  & 0.623 & 0.701 & 0.401 & 0.498 & 0.541 & 0.658  \\
				\hline
			\end{tabular}
			\caption{}
			\label{tab:personbbx}
		\end{subtable}
		\begin{subtable}[b]{.5\textwidth}
			\scriptsize
			\centering
			
			\begin{tabular}{l|cc|cc}
				\hline
				Method & \multicolumn{2}{c|}{mAP} & \multicolumn{2}{c}{mAR} \\
				\cline{2-5}
				&    medium  & large     & medium    & large   \\
				\hline
				PAF~\cite{cao2017realtime}     &   0.100  &  0.220	 & 0.113 &  0.284  \\
				SN~\cite{hidalgo2019single}     &   0.117  &  0.252	 & 0.132 &  0.315  \\
				AE~\cite{newell2017associative}        &       0.190  &  0.401   & 0.241  &  0.499\\
				OpenPose~\cite{cao2018openpose}  &       0.398 & 0.302  &    0.425  &  0.484  \\
				HRNet~\cite{sun2019deep}     &     0.471  & 0.410 &    0.538 &  0.497 \\  \hline
				Ours & \textbf{0.594} &   \textbf{0.519} &  \textbf{0.677}   & \textbf{0.635}  \\
				\hline
			\end{tabular}
			\caption{}
			\label{tab:scale}
		\end{subtable}
		\label{tab:face_and_hand}
	\end{table}

	\textbf{Effect of the person scale on whole-body pose estimation.}
	As shown in Table~\ref{tab:scale}, we investigate the effect of person scales. Interestingly, for bottom-up whole-body methods (PAF, SN and AE), the mAP for medium scale is worse than that of large scale, since they are more sensitive to the scale variance and are difficult in detecting smaller-scale people. For top-down approaches such as HRNet and ZoomNet, mAP for medium scale is better, since larger-scale person requires relatively more accurate keypoint localization. For ZoomNet, the gap between the medium and large person scale is about $7.5\%$ mAP and $4.2\%$ mAR.

	\textbf{Effect of blurriness and poses on facial landmark detection.} In Table.~\ref{tab:face_hand}, we evaluate the performance on different levels of image blurriness and facial poses (yaw angles) on WBF. The model is significantly affected by image blur (2.51 vs 19.13), while more robust to different face poses (9.02 vs 13.77).
	
	\textbf{Effect of hand poses on hand keypoint estimation.} As shown in Table.~\ref{tab:face_hand}, we evaluate the performance on different hand poses (fist, palm or others) on WBH (NME). We show that estimating the poses of ``palm'' or ``others'' (with various gestures) is more challenging than that of ``fist'' (with similar patterns). 
	
	\begin{table}[t]
		\def\arraystretch{1.2}
		\caption{\emph{left:} Effect of blurriness/poses on facial landmark detection (NME) on WholeBody-Face (WBF). \emph{right:} Effect of hand poses on hand keypoint estimation (NME) on WholeBody-Hand (WBH).}
		\scriptsize
		\begin{center}
			\begin{tabular}{c|c|c|c|c|c|c|c|c|c|c|c|c|c}
				\hline
				\multicolumn{10}{c|}{WBF (NME $\downarrow$)} &\multicolumn{4}{c}{WBH (NME $\downarrow$)} \\
				\hline
				\multicolumn{5}{c|}{Blurriness}  & \multicolumn{5}{c|}{Yaw Angles}  & \multicolumn{4}{|c}{Pose} \\ 
				\cline{1-14}
				$<1$ & $1-2$ & $2-3$ & $>3$ & ALL & $<15^{\circ}$ & $15^{\circ}-30^{\circ}$ & $30^{\circ}-45^{\circ}$ & $>45^{\circ}$ & ALL &  fist  & palm & others & ALL\\ 
				\hline
				$19.13$ & $10.85$ & $4.91$ & $2.51$ & $10.17$ & $9.02$ & $10.56$ & $12.10$ & $13.77$ & $10.17$  & $6.09$   & $7.10$ & $6.33$ & $6.66$ \\
				\hline
			\end{tabular}
		\end{center}
		\label{tab:face_hand}
	\end{table}
	
	\section{Conclusion}
	\label{sec:conclusion}
	In this paper, we proposed the first large-scale benchmark for whole-body human pose estimation. We extensively evaluate the performance of the existing approaches on our proposed COCO-WholeBody Dataset. Cross-dataset evaluation also demonstrates the generalization ability of the proposed dataset. Moreover, to solve the problem of extreme scale difference among body parts, ZoomNet is proposed to pay more attention to the hard-to-detect face/hand keypoints. Experiments show that ZoomNet significantly outperforms the prior arts. 
	
	~\\
	\textbf{Acknowledgement.} This work is partially supported by the SenseTime Donation for Research, HKU Seed Fund for Basic Research, Startup Fund, General Research Fund No.27208720, the Australian Research Council Grant DP200103223 and Australian Medical Research Future Fund MRFAI000085.
	
	%
	%
	\bibliographystyle{splncs04}
	\bibliography{egbib}

\begin{thebibliography}{10}
\providecommand{\url}[1]{\texttt{#1}}
\providecommand{\urlprefix}{URL }
\providecommand{\doi}[1]{https://doi.org/#1}

\bibitem{alp2018densepose}
Alp~G{\"u}ler, R., Neverova, N., Kokkinos, I.: Densepose: Dense human pose
  estimation in the wild. In: Proceedings of the IEEE Conference on Computer
  Vision and Pattern Recognition (CVPR) (2018)

\bibitem{andriluka2018posetrack}
Andriluka, M., Iqbal, U., Insafutdinov, E., Pishchulin, L., Milan, A., Gall,
  J., Schiele, B.: Posetrack: A benchmark for human pose estimation and
  tracking. In: Proceedings of the IEEE Conference on Computer Vision and
  Pattern Recognition (CVPR) (2018)

\bibitem{andriluka14cvpr}
Andriluka, M., Pishchulin, L., Gehler, P., Schiele, B.: 2d human pose
  estimation: New benchmark and state of the art analysis. In: Proceedings of
  the IEEE Conference on Computer Vision and Pattern Recognition (CVPR) (2014)

\bibitem{belhumeur2013localizing}
Belhumeur, P.N., Jacobs, D.W., Kriegman, D.J., Kumar, N.: Localizing parts of
  faces using a consensus of exemplars. IEEE transactions on pattern analysis
  and machine intelligence  (2013)

\bibitem{Burgos2013Robust}
Burgos-Artizzu, X.P., Perona, P., Dollár, P.: Robust face landmark estimation
  under occlusion. In: Proceedings of the 2013 IEEE International Conference on
  Computer Vision (2013)

\bibitem{cao2014face}
Cao, X., Wei, Y., Wen, F., Sun, J.: Face alignment by explicit shape
  regression. International Journal of Computer Vision  (2014)

\bibitem{cao2018openpose}
Cao, Z., Hidalgo, G., Simon, T., Wei, S.E., Sheikh, Y.: Openpose: realtime
  multi-person 2d pose estimation using part affinity fields. arXiv preprint
  arXiv:1812.08008  (2018)

\bibitem{cao2017realtime}
Cao, Z., Simon, T., Wei, S.E., Sheikh, Y.: Realtime multi-person 2d pose
  estimation using part affinity fields. In: Proceedings of the IEEE Conference
  on Computer Vision and Pattern Recognition (CVPR) (2017)

\bibitem{chen2018cascaded}
Chen, Y., Wang, Z., Peng, Y., Zhang, Z., Yu, G., Sun, J.: Cascaded pyramid
  network for multi-person pose estimation. In: Proceedings of the IEEE
  Conference on Computer Vision and Pattern Recognition (CVPR) (2018)

\bibitem{duan2019trb}
Duan, H., Lin, K.Y., Jin, S., Liu, W., Qian, C., Ouyang, W.: Trb: A novel
  triplet representation for understanding 2d human body. In: Proceedings of
  the IEEE International Conference on Computer Vision. pp. 9479--9488 (2019)

\bibitem{eichner2010we}
Eichner, M., Ferrari, V.: We are family: Joint pose estimation of multiple
  persons. In: Proceedings of the European Conference on Computer Vision (ECCV)
  (2010)

\bibitem{fang2017rmpe}
Fang, H.S., Xie, S., Tai, Y.W., Lu, C.: Rmpe: Regional multi-person pose
  estimation. In: Proceedings of the IEEE Conference on Computer Vision and
  Pattern Recognition (CVPR) (2017)

\bibitem{gomez2017large}
Gomez-Donoso, F., Orts-Escolano, S., Cazorla, M.: Large-scale multiview 3d hand
  pose dataset. arXiv preprint arXiv:1707.03742  (2017)

\bibitem{Gross2010Image}
Gross, R., Matthews, I., Cohn, J., Kanade, T., Baker, S.: Multi-pie. In: Image
  and Vision Computing (2010)

\bibitem{guan2006multi}
Guan, H., Chang, J.S., Chen, L., Feris, R.S., Turk, M.: Multi-view
  appearance-based 3d hand pose estimation. In: IEEE Conference on Computer
  Vision and Pattern Recognition Workshop (2006)

\bibitem{he2017mask}
He, K., Gkioxari, G., Doll{\'a}r, P., Girshick, R.: Mask r-cnn. arXiv preprint
  arXiv:1703.06870  (2017)

\bibitem{hidalgo2019single}
Hidalgo, G., Raaj, Y., Idrees, H., Xiang, D., Joo, H., Simon, T., Sheikh, Y.:
  Single-network whole-body pose estimation. In: Proceedings of the IEEE
  Conference on Computer Vision and Pattern Recognition (CVPR) (2019)

\bibitem{honari2016recombinator}
Honari, S., Yosinski, J., Vincent, P., Pal, C.: Recombinator networks: Learning
  coarse-to-fine feature aggregation. In: Proceedings of the IEEE Conference on
  Computer Vision and Pattern Recognition (CVPR) (2016)

\bibitem{Insafutdinov2016ArtTrack}
Insafutdinov, E., Andriluka, M., Pishchulin, L., Tang, S., Levinkov, E.,
  Andres, B., Schiele, B., Campus, S.I.: Arttrack: Articulated multi-person
  tracking in the wild. In: Proceedings of the IEEE Conference on Computer
  Vision and Pattern Recognition (CVPR) (2017)

\bibitem{Insafutdinov2016DeeperCut}
Insafutdinov, E., Pishchulin, L., Andres, B., Andriluka, M., Schiele, B.:
  Deepercut: A deeper, stronger, and faster multi-person pose estimation model.
  In: Proceedings of the European Conference on Computer Vision (ECCV) (2016)

\bibitem{Iqbal2016PoseTrack}
Iqbal, U., Milan, A., Gall, J.: Pose-track: Joint multi-person pose estimation
  and tracking. arXiv preprint arXiv:1611.07727  (2016)

\bibitem{jin2019multi}
Jin, S., Liu, W., Ouyang, W., Qian, C.: Multi-person articulated tracking with
  spatial and temporal embeddings. In: Proceedings of the IEEE Conference on
  Computer Vision and Pattern Recognition (CVPR) (2019)

\bibitem{jin2017towards}
Jin, S., Ma, X., Han, Z., Wu, Y., Yang, W., Liu, W., Qian, C., Ouyang, W.:
  Towards multi-person pose tracking: Bottom-up and top-down methods. In: IEEE
  International Conference on Computer Vision Workshop (2017)

\bibitem{joo2018total}
Joo, H., Simon, T., Sheikh, Y.: Total capture: A 3d deformation model for
  tracking faces, hands, and bodies. In: Proceedings of the IEEE Conference on
  Computer Vision and Pattern Recognition (CVPR) (2018)

\bibitem{kingma2014adam}
Kingma, D.P., Ba, J.: Adam: A method for stochastic optimization. arXiv
  preprint arXiv:1412.6980  (2014)

\bibitem{koestinger2011annotated}
Koestinger, M., Wohlhart, P., Roth, P.M., Bischof, H.: Annotated facial
  landmarks in the wild: A large-scale, real-world database for facial landmark
  localization. In: IEEE International Conference on Computer Vision Workshop
  (2011)

\bibitem{kowalski2017deep}
Kowalski, M., Naruniec, J., Trzcinski, T.: Deep alignment network: A
  convolutional neural network for robust face alignment. In: IEEE Conference
  on Computer Vision and Pattern Recognition Workshop (2017)

\bibitem{law2018cornernet}
Law, H., Deng, J.: Cornernet: Detecting objects as paired keypoints. In:
  Proceedings of the European Conference on Computer Vision (ECCV) (2018)

\bibitem{le2012interactive}
Le, V., Brandt, J., Lin, Z., Bourdev, L., Huang, T.S.: Interactive facial
  feature localization. In: Proceedings of the European Conference on Computer
  Vision (ECCV) (2012)

\bibitem{li2019crowdpose}
Li, J., Wang, C., Zhu, H., Mao, Y., Fang, H.S., Lu, C.: Crowdpose: Efficient
  crowded scenes pose estimation and a new benchmark. In: Proceedings of the
  IEEE Conference on Computer Vision and Pattern Recognition. pp. 10863--10872
  (2019)

\bibitem{lin2014microsoft}
Lin, T.Y., Maire, M., Belongie, S., Hays, J., Perona, P., Ramanan, D.,
  Doll{\'a}r, P., Zitnick, C.L.: Microsoft coco: Common objects in context. In:
  Proceedings of the European Conference on Computer Vision (ECCV) (2014)

\bibitem{liu2018cascaded}
Liu, W., Chen, J., Li, C., Qian, C., Chu, X., Hu, X.: A cascaded inception of
  inception network with attention modulated feature fusion for human pose
  estimation. In: The Thirty-Second AAAI Conference on Artificial Intelligence
  (2018)

\bibitem{messer1999xm2vtsdb}
Messer, K., Matas, J., Kittler, J., Luettin, J., Maitre, G.: Xm2vtsdb: The
  extended m2vts database. In: Second international conference on audio and
  video-based biometric person authentication (1999)

\bibitem{mueller2018ganerated}
Mueller, F., Bernard, F., Sotnychenko, O., Mehta, D., Sridhar, S., Casas, D.,
  Theobalt, C.: Ganerated hands for real-time 3d hand tracking from monocular
  rgb. In: Proceedings of the IEEE Conference on Computer Vision and Pattern
  Recognition (CVPR) (2018)

\bibitem{mueller2017real}
Mueller, F., Mehta, D., Sotnychenko, O., Sridhar, S., Casas, D., Theobalt, C.:
  Real-time hand tracking under occlusion from an egocentric rgb-d sensor. In:
  Proceedings of International Conference on Computer Vision ({ICCV}) (2017)

\bibitem{neverova2014multi}
Neverova, N., Wolf, C., Taylor, G.W., Nebout, F.: Multi-scale deep learning for
  gesture detection and localization. In: Proceedings of the European
  Conference on Computer Vision (ECCV) (2014)

\bibitem{newell2017associative}
Newell, A., Huang, Z., Deng, J.: Associative embedding: End-to-end learning for
  joint detection and grouping. In: Advances in Neural Information Processing
  Systems (2017)

\bibitem{newell2016stacked}
Newell, A., Yang, K., Deng, J.: Stacked hourglass networks for human pose
  estimation. In: Proceedings of the European Conference on Computer Vision
  (ECCV) (2016)

\bibitem{nie2017generative}
Nie, X., Feng, J., Xing, J., Yan, S.: Generative partition networks for
  multi-person pose estimation. arXiv preprint arXiv:1705.07422  (2017)

\bibitem{oikonomidis2012tracking}
Oikonomidis, I., Kyriazis, N., Argyros, A.A.: Tracking the articulated motion
  of two strongly interacting hands. In: IEEE Conference on Computer Vision and
  Pattern Recognition (2012)

\bibitem{papandreou2018personlab}
Papandreou, G., Zhu, T., Chen, L.C., Gidaris, S., Tompson, J., Murphy, K.:
  Personlab: Person pose estimation and instance segmentation with a bottom-up,
  part-based, geometric embedding model. arXiv preprint arXiv:1803.08225
  (2018)

\bibitem{papandreou2017towards}
Papandreou, G., Zhu, T., Kanazawa, N., Toshev, A., Tompson, J., Bregler, C.,
  Murphy, K.: Towards accurate multi-person pose estimation in the wild. arXiv
  preprint arXiv:1701.01779  (2017)

\bibitem{Pech2000Diatom}
Pech-Pacheco, J.~L., C., G., Chamorro-Martinez, J., Fernández-Valdivia, J.:
  Diatom autofocusing in brightfield microscopy: a comparative study. In: ICPR
  (2000)

\bibitem{phillips2005overview}
Phillips, P.J., Flynn, P.J., Scruggs, T., Bowyer, K.W., Chang, J., Hoffman, K.,
  Marques, J., Min, J., Worek, W.: Overview of the face recognition grand
  challenge. In: Proceedings of the IEEE Conference on Computer Vision and
  Pattern Recognition (CVPR) (2005)

\bibitem{pishchulin2016deepcut}
Pishchulin, L., Insafutdinov, E., Tang, S., Andres, B., Andriluka, M., Gehler,
  P.V., Schiele, B.: Deepcut: Joint subset partition and labeling for multi
  person pose estimation. In: Proceedings of the IEEE Conference on Computer
  Vision and Pattern Recognition (CVPR) (2016)

\bibitem{renNIPS15fasterrcnn}
Ren, S., He, K., Girshick, R., Sun, J.: Faster {R-CNN}: Towards real-time
  object detection with region proposal networks. In: Advances in Neural
  Information Processing Systems (NIPS) (2015)

\bibitem{romero2017embodied}
Romero, J., Tzionas, D., Black, M.J.: Embodied hands: Modeling and capturing
  hands and bodies together. ACM Transactions on Graphics (ToG)  (2017)

\bibitem{Ronchi2017Benchmarking}
Ronchi, M.R., Perona, P.: {Benchmarking and Error Diagnosis in Multi-Instance
  Pose Estimation}. Proceedings of International Conference on Computer Vision
  ({ICCV})  (2017)

\bibitem{sagonas2013300}
Sagonas, C., Tzimiropoulos, G., Zafeiriou, S., Pantic, M.: 300 faces
  in-the-wild challenge: The first facial landmark localization challenge. In:
  IEEE International Conference on Computer Vision Workshop (2013)

\bibitem{sharp2015accurate}
Sharp, T., Keskin, C., Robertson, D., Taylor, J., Shotton, J., Kim, D.,
  Rhemann, C., Leichter, I., Vinnikov, A., Wei, Y., et~al.: Accurate, robust,
  and flexible real-time hand tracking. In: Proceedings of the 33rd Annual ACM
  Conference on Human Factors in Computing Systems (2015)

\bibitem{simon2017hand}
Simon, T., Joo, H., Matthews, I., Sheikh, Y.: Hand keypoint detection in single
  images using multiview bootstrapping. In: Proceedings of the IEEE Conference
  on Computer Vision and Pattern Recognition (CVPR) (2017)

\bibitem{sridhar2015fast}
Sridhar, S., Mueller, F., Oulasvirta, A., Theobalt, C.: Fast and robust hand
  tracking using detection-guided optimization. In: Proceedings of the IEEE
  Conference on Computer Vision and Pattern Recognition (CVPR) (2015)

\bibitem{sun2019deep}
Sun, K., Xiao, B., Liu, D., Wang, J.: Deep high-resolution representation
  learning for human pose estimation. arXiv preprint arXiv:1902.09212  (2019)

\bibitem{sun2019high}
Sun, K., Zhao, Y., Jiang, B., Cheng, T., Xiao, B., Liu, D., Mu, Y., Wang, X.,
  Liu, W., Wang, J.: High-resolution representations for labeling pixels and
  regions. arXiv preprint arXiv:1904.04514  (2019)

\bibitem{sun2013deep}
Sun, Y., Wang, X., Tang, X.: Deep convolutional network cascade for facial
  point detection. In: Proceedings of the IEEE Conference on Computer Vision
  and Pattern Recognition (CVPR) (2013)

\bibitem{tompson2014real}
Tompson, J., Stein, M., Lecun, Y., Perlin, K.: Real-time continuous pose
  recovery of human hands using convolutional networks. ACM Transactions on
  Graphics (ToG)  (2014)

\bibitem{trigeorgis2016mnemonic}
Trigeorgis, G., Snape, P., Nicolaou, M.A., Antonakos, E., Zafeiriou, S.:
  Mnemonic descent method: A recurrent process applied for end-to-end face
  alignment. In: Proceedings of the IEEE Conference on Computer Vision and
  Pattern Recognition (CVPR) (2016)

\bibitem{tzimiropoulos2015project}
Tzimiropoulos, G.: Project-out cascaded regression with an application to face
  alignment. In: Proceedings of the IEEE Conference on Computer Vision and
  Pattern Recognition (CVPR) (2015)

\bibitem{valle2018deeply}
Valle, R., Buenaposada, J.M., Valdes, A., Baumela, L.: A deeply-initialized
  coarse-to-fine ensemble of regression trees for face alignment. In:
  Proceedings of the European Conference on Computer Vision (ECCV) (2018)

\bibitem{Yangang2018Mask}
Wang, Y., Peng, C., Liu, Y.: Mask-pose cascaded cnn for 2d hand pose estimation
  from single color image. IEEE Transactions on Circuits and Systems for Video
  Technology  (2018)

\bibitem{wei2016convolutional}
Wei, S.E., Ramakrishna, V., Kanade, T., Sheikh, Y.: Convolutional pose
  machines. In: The IEEE Conference on Computer Vision and Pattern Recognition
  (CVPR) (2016)

\bibitem{wu2017ai}
Wu, J., Zheng, H., Zhao, B., Li, Y., Yan, B., Liang, R., Wang, W., Zhou, S.,
  Lin, G., Fu, Y., et~al.: Ai challenger: a large-scale dataset for going
  deeper in image understanding. arXiv preprint arXiv:1711.06475  (2017)

\bibitem{wu2018look}
Wu, W., Qian, C., Yang, S., Wang, Q., Cai, Y., Zhou, Q.: Look at boundary: A
  boundary-aware face alignment algorithm. In: Proceedings of the IEEE
  Conference on Computer Vision and Pattern Recognition (CVPR) (2018)

\bibitem{xiang2019monocular}
Xiang, D., Joo, H., Sheikh, Y.: Monocular total capture: Posing face, body, and
  hands in the wild. In: Proceedings of the IEEE Conference on Computer Vision
  and Pattern Recognition (CVPR) (2019)

\bibitem{xiao2018simple}
Xiao, B., Wu, H., Wei, Y.: Simple baselines for human pose estimation and
  tracking. In: Proceedings of the European Conference on Computer Vision
  (ECCV) (2018)

\bibitem{xiong2013supervised}
Xiong, X., De~la Torre, F.: Supervised descent method and its applications to
  face alignment. In: Proceedings of the IEEE Conference on Computer Vision and
  Pattern Recognition (CVPR) (2013)

\bibitem{yuan2017bighand2}
Yuan, S., Ye, Q., Stenger, B., Jain, S., Kim, T.K.: Bighand2. 2m benchmark:
  Hand pose dataset and state of the art analysis. In: Proceedings of the IEEE
  Conference on Computer Vision and Pattern Recognition (CVPR) (2017)

\bibitem{zhang2015learning}
Zhang, Z., Luo, P., Loy, C.C., Tang, X.: Learning deep representation for face
  alignment with auxiliary attributes. IEEE transactions on pattern analysis
  and machine intelligence  (2015)

\bibitem{zhu2012face}
Zhu, X., Ramanan, D.: Face detection, pose estimation, and landmark
  localization in the wild. In: Proceedings of the IEEE Conference on Computer
  Vision and Pattern Recognition (CVPR) (2012)

\bibitem{zb2017hand}
Zimmermann, C., Brox, T.: Learning to estimate 3d hand pose from single rgb
  images. arXiv preprint arXiv: 1705.01389  (2017)

\bibitem{Freihand2019}
Zimmermann, C., Ceylan, D., Yang, J., Russell, B., Argus, M., Brox, T.:
  Freihand: A dataset for markerless capture of hand pose and shape from single
  rgb images. In: Proceedings of International Conference on Computer Vision
  ({ICCV}) (2019)

\end{thebibliography}
	
	\appendix
	
	\clearpage
	
	\section{Annotation Details}
	
	The annotation of face/hand keypoints in our COCO-WholeBody dataset follows semi-automatic methodology. Firstly, face/hand bounding boxes are annotated manually. Secondly, we utilize a face model and a hand model, which are trained on large-scale face datasets and hand datasets respectively, to pre-annotate the face and hand keypoints. Next, manual correction of the face/hand keypoints is conducted. Foot keypoints are directly manually labeled. Note that, quality inspections are conducted in every step.
	
\begin{figure}[tb]
		\centering
		\includegraphics[width=0.7\textwidth]{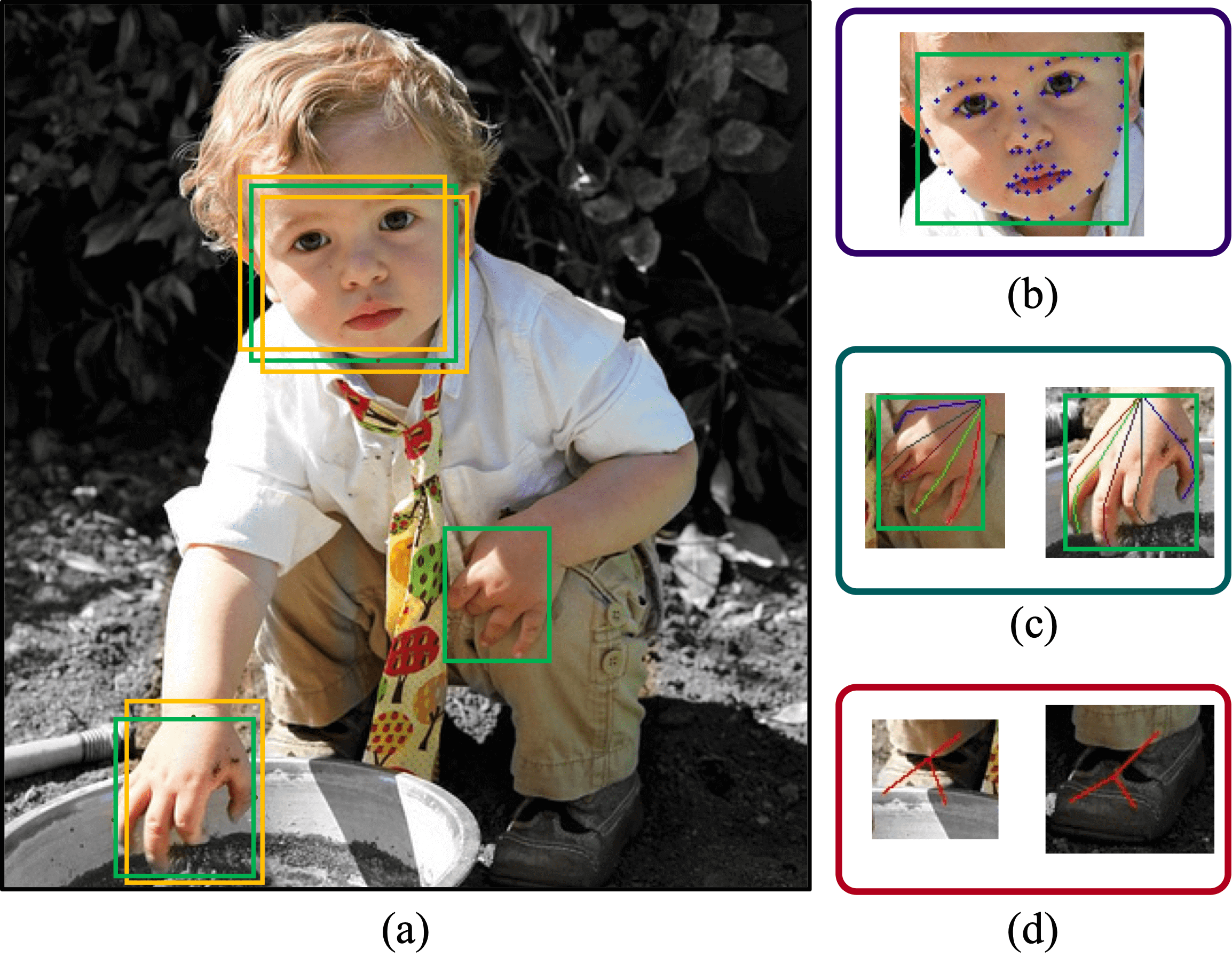}
		\caption{Face/hand bounding box annotation. Bounding boxes should tightly enclose all the keypoints. Positive (green) and negative (orange) cases are shown.}
		\label{fig:anno_box}
	\end{figure}

\begin{figure*}[tb]
		\centering
		\includegraphics[width=0.95\textwidth]{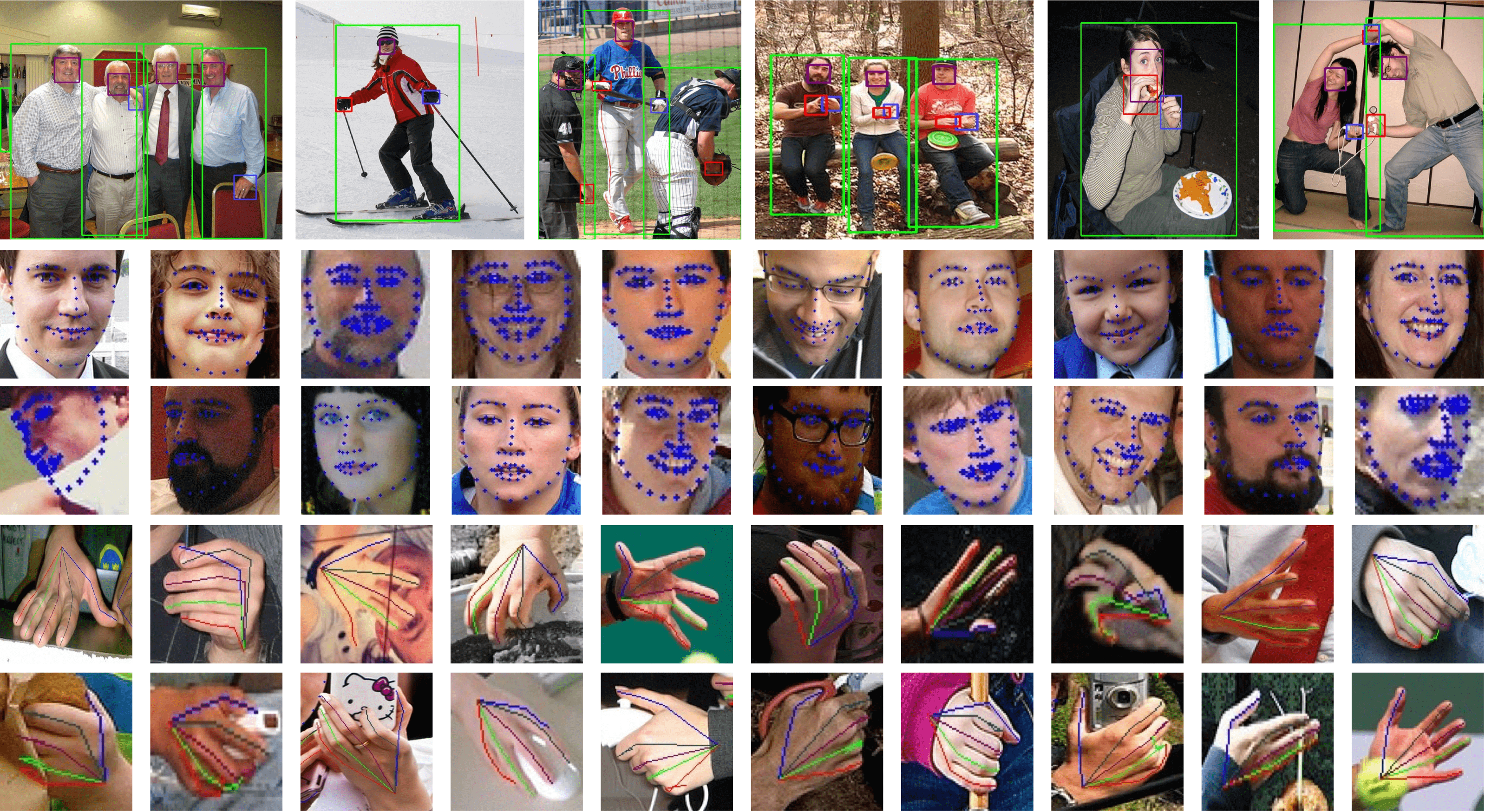}
		\caption{Annotation examples. Line \#1: We use different colors to distinguish different types of bounding boxes, \ie body (green), face (purple), left hand (blue) and right hand (red). Line \#2 and Line\#3: Face keypoints. Line \#4 and Line\#5: Hand keypoints.}
		\label{fig:anno_example}
	\end{figure*}

	\textbf{Face and hand bounding box:}
	To ensure the quality of face/hand bounding boxes, well-defined standards are followed. \textbf{Face bounding box} is labeled only if the box is bigger than 8 pixels and the rotation angle of the face is less than $100^{\circ}$ from the frontal view. As for some special cases, faces of real persons in photos, posters, and clothes are labeled but faces of sculptures, models, cartoons, paintings, and animals are not. The face bounding box is defined as the minimal bounding rectangle of the face keypoints. Quality inspections are conducted by another group of annotators and bounding boxes whose positions are inaccurate are re-annotated. \textbf{Hand bounding box} is labeled when the hand image is vivid and the position of the hand keypoints can be well-determined. The box is regarded as invalid if the corresponding hand is severely occluded or part of the hand is out of the image. Special case settings follow those of face bounding box and independent quality inspections are conducted. Examples of face/hand bounding boxes are shown in Fig.~\ref{fig:anno_box}, where only the green boxes meet our annotation requirements. More visualization results for bounding boxes are demonstrated in Fig.~\ref{fig:anno_example} Line\#1. We have three types of bounding boxes, \ie body (green), face (purple), left hand (blue) and right hand (red). 
	
	\textbf{Face Keypoints:}
	We apply the 68-joint face model \cite{sagonas2013300} as shown in Fig.~\ref{fig:anno_box}(b). A few occluded keypoints may be estimated by annotators if most keypoints are visible in the image. In Fig.~\ref{fig:anno_example}, Line\#2 and Line\#3 visualize more examples of the face keypoint annotations.

	\textbf{Hand Keypoints:}
	Self-occlusion is very common for hand keypoints. As a result, the annotation for hand keypoints requires trained experts and enormous workload although pseudo labels are given. We use 21-joint hand model \cite{simon2017hand} and annotate quite a lot of challenging cases. Annotation is shown in Fig.~\ref{fig:anno_box}(c) and more examples are visualized in Fig.~\ref{fig:anno_example}, where Line\#4 and Line\#5 visualize some examples of the hand keypoint annotations for various hand poses.
	
	\textbf{Foot Keypoints:}
	Six foot keypoints are defined following \cite{cao2018openpose}. The order in the annotation file is as follows: left big toe, left small toe, left heel, right big toe, right small toe, and right heel. The keypoints are defined in the inner center rather than on the surface to fit in images in different views. Qualitative examples are shown in Fig.~\ref{fig:anno_box}(d).

	\section{Baseline Implementation Details}
	
	We used the official codes to reproduce existing methods. We keep all training parameters (e.g. input size, \#iterations, learning rate, and so on) the same, except \#keypoints (\# means the number of). We also trained all the existing methods on the original 17-keypoint COCO dataset and verified that our re-implementation is the same as the original papers. For fair comparisons, all experimental results are obtained with single-scale testing. The implementation details of the baseline methods we used in the experiments are listed as following:
	
	\textbf{OpenPose Whole-body System}~\cite{cao2018openpose} is a Multi-Network whole-body pose estimation system, which consists of a body keypoint model, a facial landmark detector and a hand pose estimator. We reimplement the approach by training these models on COCO-WholeBody dataset separately based on the official training codes~\footnote{https://github.com/CMU-Perceptual-Computing-Lab/openpose}.
	
	\textbf{Single-Network Whole-body Pose Estimation}~\cite{hidalgo2019single} is a recently proposed method for whole-body pose estimation. We follow~\cite{hidalgo2019single} and retrain the whole-body keypoint estimator~\footnote{https://github.com/CMU-Perceptual-Computing-Lab/openpose\_train} in our COCO-WholeBody dataset. The number of keypoints is 133, and the number of PAFs is 134 as we designed a tree structure except for the two loops around the lips. Face, hand and foot keypoints are connected to the corresponding nearest body keypoints. Following~\cite{hidalgo2019single}, we applied 3 stages for PAF and 1 stage for confidence maps. We use a batch size of 10 images in each GPU and Adam optimization with an initial learning rate of 1e-3 to train the model.
	
\begin{figure*}[tb]
		\centering
		\includegraphics[width=0.65\textwidth]{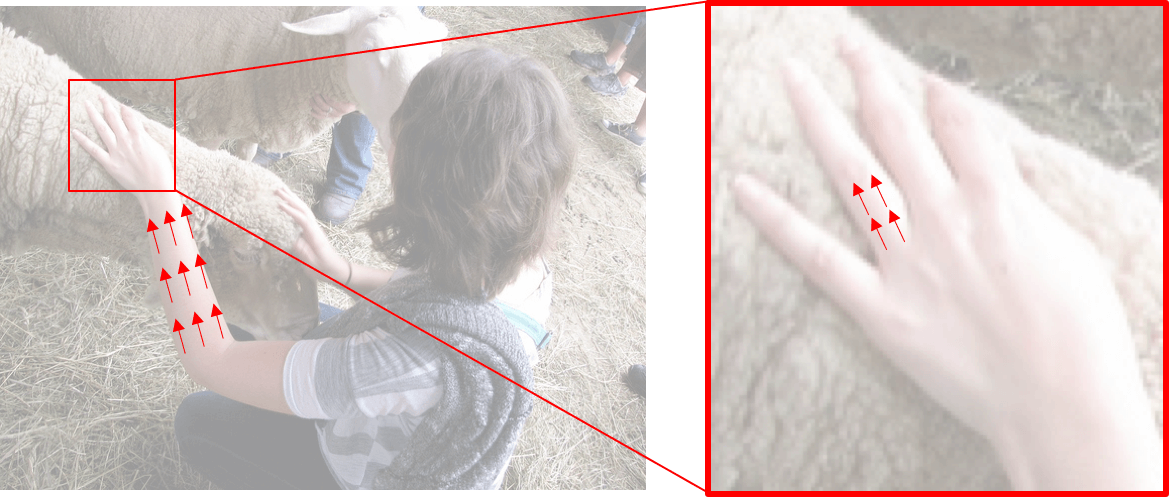}
		\caption{Visualizations of Part-affinity Fields.}
		\label{fig:paf}
	\end{figure*}

	\textbf{Part-affinity Fields (PAF)}~\cite{cao2017realtime} is also re-implemented for the whole-body pose estimation task based on the open-source codes~\footnote{https://github.com/tensorboy/pytorch\_Realtime\_Multi-Person\_Pose\_Estimation}. The settings of PAFs and confidence maps are the same as Single-Network~\cite{hidalgo2019single} and CPM~\cite{wei2016convolutional} network is used as its backbone. We use SGD with an initial learning rate of 1 to train the model. Note that, the direction of limb (or value of the affinity fields) is calculated in the image scale before down-sampling, see Fig.~\ref{fig:paf}. Therefore, for most tiny hands and faces, the PAF prediction and keypoint grouping will not be affected.

	\textbf{Associative Embedding (AE)}~\cite{newell2017associative} learns to group keypoints by associative embedding, which is flexible in terms of various numbers of keypoints to predict. The official open-source codes~\footnote{https://github.com/princeton-vl/pose-ae-train} are used in our implementation. We use the 4-stacked hourglass backbone and follow the same training settings as in~\cite{newell2017associative} in our experiments. 
	
	\textbf{HRNet}~\cite{sun2019deep} is the recent state-of-the-art model for the task of multi-person human pose estimation. We retrain the model~\footnote{https://github.com/leoxiaobin/deep-high-resolution-net.pytorch} to fit for the whole-body pose estimation task by directly adding the number of keypoints to 133. For fair comparisons, we choose HRNet-w32 as the backbone in the experiments. Note that this model can be viewed as the single-stage alternative of our multi-stage ZoomNet. The comparison between HRNet and ZoomNet demonstrates the effectiveness of the multi-stage keypoint localization. 
	
	\section{ZoomNet Implementation Details}
	
	We use 2D gaussian confidence heatmaps with $\sigma=3$ to encode the keypoint locations.
	The sum of squared error (SSE) loss function between the predicted heatmaps and the ground truth heatmaps is used for training both corner keypoints and body keypoints. The losses of different body parts (body, face, hand, and feet) are summed up with the same loss weight.
	
	We follow the same setting as HRNet~\cite{sun2019deep} to use data augmentation with random scaling ([-35\%, 35\%]), random rotation ([$-45^{\circ}$, $45^{\circ}$]) and flipping. 
	BodyNet and FaceHead/HandHead are first pre-trained separately and then end-to-end finetuned as a whole for 120 epochs in total. ZoomNet is trained on 8 GPUs with a batch size of 32 in each GPU. We use Adam~\cite{kingma2014adam} with the base learning rate of 1e-3, and decay it to 1e-4 and 1e-5 at the $80$th and $100$th epochs respectively.

	\section{Analysis}
	
	\subsubsection{Experiments on Foot Keypoint Dataset}
	Cao \etal released the first human foot dataset~\cite{cao2018openpose} (COCO-foot), which extends COCO~\cite{lin2014microsoft} dataset with 15k foot annotations. We also evaluate our proposed ZoomNet on COCO-foot dataset and directly compare with OpenPose~\cite{cao2018openpose} and SN~\cite{hidalgo2019single} in Table~\ref{tab:coco_foot}. We find that our proposed ZoomNet outperforms SN significantly.
	
	\begin{table}[t]
		\caption{Body-foot AP on COCO-foot benchmark~\cite{cao2018openpose}. Some results are copied from~\cite{hidalgo2019single}. Our proposed ZoomNet outperforms SN significantly.}
		\begin{center}
			\begin{tabular}{l|c|c}
				\hline
				Method    & Body AP  &  Foot AP \\
				\hline
				Body-foot OpenPose (multi-scale)~\cite{cao2018openpose} &  65.3 & 77.9 \\
				Body-foot SN (multi-scale)~\cite{hidalgo2019single} &  66.4 & 76.8 \\
				Body-foot ZoomNet &  \textbf{75.4} & \textbf{84.7} \\
				\hline
			\end{tabular}
		\end{center}
		\label{tab:coco_foot}
	\end{table}
	
	\subsection{Experiments about joint learning.}
	
	In Table~\ref{tab:joint_learning}, we explore the effectiveness of joint training of BodyNet, FaceHead and HandHead in ZoomNet. We compare (1) joint training, (2) reusing features, and (3) fully independent face/hand detectors. Joint learning improves over ``reusing features'' on the performance of face (0.623 vs 0.609) and hand (0.401 vs 0.393) for more efficient feature learning. Fully independent method requires two additional models with increased complexity, but achieves limited gain (0.543 vs 0.541).
	
	\begin{table}[t]
		\caption{Effectiveness of joint learning.}
		\begin{center}
			\begin{tabular}{l|c|c|c|c|c}
				\hline
				Method & Body AP & Foot AP & Face AP & Hand AP & WholeBody AP \\
				\hline
				joint training	& 0.743 &	0.798 &	0.623 &	0.401 &	0.541 \\
				reusing features &	0.745 &	0.796 &	0.609 &	0.393 &	0.539 \\
				fully independent &	0.745 &	0.796 &	0.623 &	0.419 &	0.543 \\
				\hline
			\end{tabular}
		\end{center}
		\label{tab:joint_learning}
	\end{table}

	\subsubsection{Face/Hand Bounding Box Detection}
	
	In this section, we compare the results of face and hand bounding box detection. Compared to human body detection, detecting small objects such as face and hands are more challenging, since they only occupy a relatively small area in the whole image. General detection approaches such as Faster RCNN~\cite{renNIPS15fasterrcnn} usually treat body/face/hands as normal objects and detect all of them at once. However, note that the human body is inherently a multi-level structure, where the face/hands are low-level objects of the high-level human body. Intuitively, the location of the human body will guide the detection of face/hands. Common detection methods usually ignore the inherent correlation between the human body and the face/hands, which will lead to inferior performance. To deal with the scale variance problem, ZoomNet first locates all the person bounding boxes from the image and then detects the face and hands in each bounding box. This multi-level design enables the model to focus on the potential location of the sub-objects and ignore the disturbing background. Therefore, it is beneficial for detecting small sub-objects such as face and hands. As shown in Table~\ref{tab:detection}, ZoomNet outperforms the Faster RCNN model by a large margin, demonstrating the effectiveness of our multi-level object detection.

	\begin{table}[t]
		\caption{Face/hand bounding box detection results on our COCO-WholeBody benchmark. Our proposed ZoomNet outperforms Faster RCNN~\cite{renNIPS15fasterrcnn} because of its multi-level design which better handles the scale variance.} 
		\begin{center}
			\begin{tabular}{l|cc|cc|cc}
				\hline
				Method & \multicolumn{2}{c|}{face}  & \multicolumn{2}{c|}{lefthand}  &  \multicolumn{2}{c}{righthand} \\
				\cline{2-7}
				&  AP     & AR     & AP   & AR     &  AP  & AR\\
				\hline
				Faster RCNN~\cite{renNIPS15fasterrcnn} & 0.439 & 0.712 & 0.266 & 0.440 & 0.262 & 0.430 \\
				ZoomNet & \textbf{0.582} & \textbf{0.728} & \textbf{0.349} & \textbf{0.463} & \textbf{0.356} & \textbf{0.458}  \\
				\hline
			\end{tabular}
		\end{center}
		\label{tab:detection}
	\end{table}

	\subsubsection{Error Analysis}
	
\begin{figure*}[tb]
		\centering
		\includegraphics[width=0.99\textwidth]{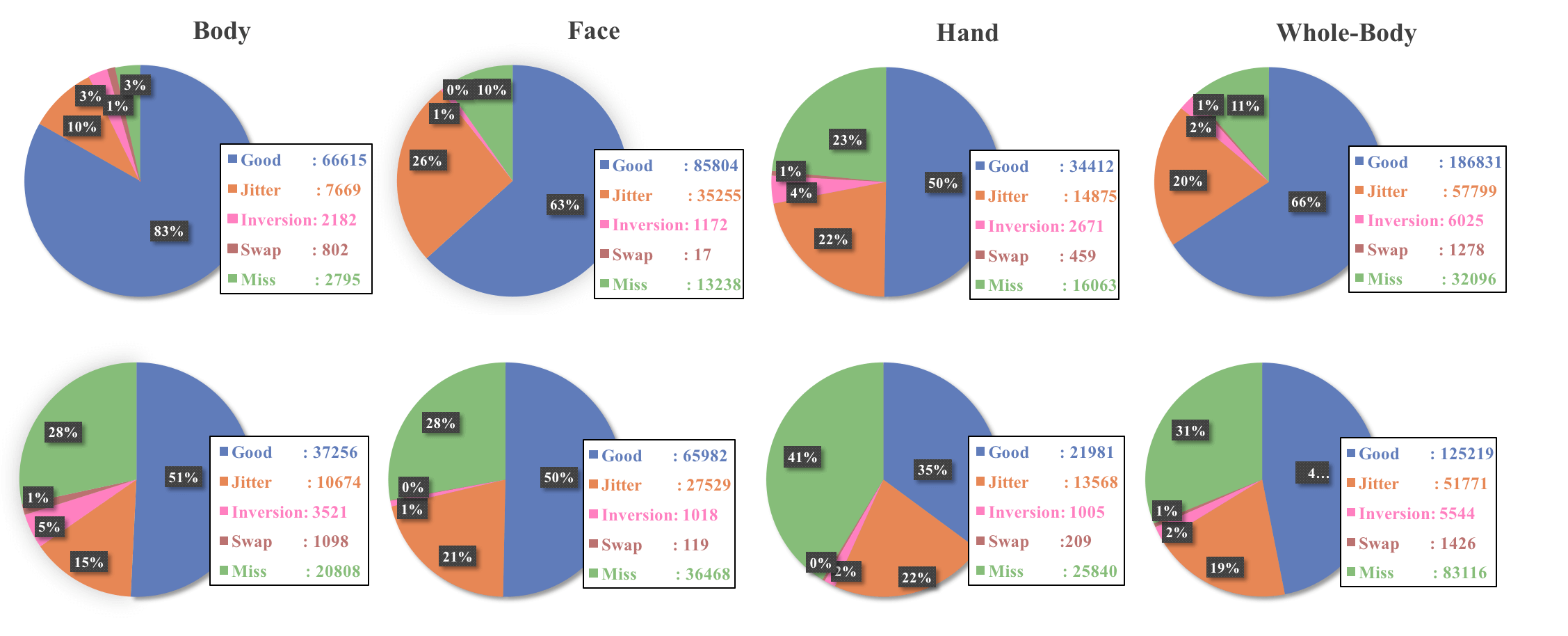}
		\caption{Localization error comparison between our proposed ZoomNet (top) and Single-Network~\cite{hidalgo2019single} (bottom). ZoomNet significantly outperforms Single-Network in the distribution of the localization error for body, face, hand and whole-body.}
		\label{fig:error}
	\end{figure*}

	In this section, we provide a more detailed error analysis for ZoomNet and Single-Network~\cite{hidalgo2019single}. The breakdown of errors over different body parts is shown in Fig.~\ref{fig:error}. We follow~\cite{Ronchi2017Benchmarking} to define four types of localization errors, \ie Jitter, Miss, Inversion, and Swap. \emph{Jitter} means small error around the correct keypoint location, while \emph{Miss} means the detection is not within the proximity of any ground truth body part. \emph{Inversion} means the joint type of detected keypoint is wrong. \emph{Swap} means the detected keypoint is grouped to a wrong person instance. On the other hand, \emph{Good} indicates correct prediction. 
	
	We use the pie chart to show the distribution of the localization errors for the body, face, hand, and whole-body. \emph{Miss} is the major error for all parts, and the accuracy of the hand keypoints is lower than that of the body and face keypoints. Also, ZoomNet has a higher proportion of \emph{Good} keypoints than Single-Network.
	
	\subsubsection{Size Sensitivity}
\begin{figure*}[tb]
		\centering
		\includegraphics[width=0.99\textwidth]{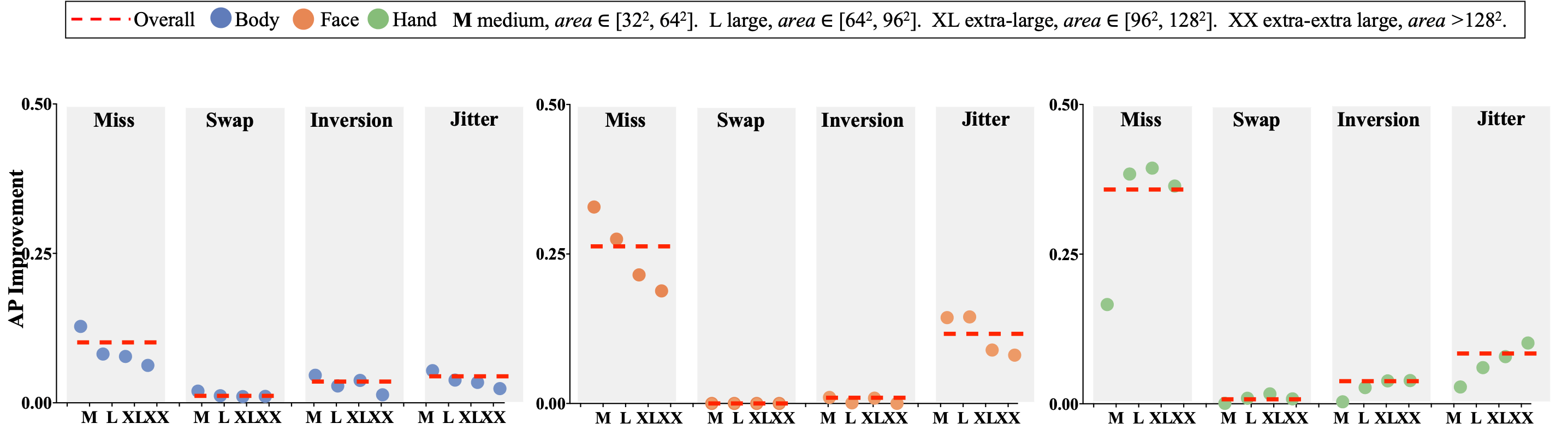}
		\caption{The AP improvement obtained by correcting each type of error (including Miss, Swap, Inversion, and Jitter) for body, face, and hand separately. We use the dashed red lines to indicate performance improvement over all the instance sizes.}
		\label{fig:size}
	\end{figure*}

	In this section, we analyze the sensitivity of our proposed ZoomNet to different person sizes. To this end, we separate the COCO-WholeBody dataset into four size groups: \ie medium (M), large (L), extra-large (XL) and extra-extra large (XX). We follow~\cite{Ronchi2017Benchmarking} to use the area of the person to measure the person size, M for $area \in [32^2, 64^2]$, L for $area \in [64^2, 96^2]$, XL for $area \in [96^2, 128^2]$, and XX for $area > 128^2$. In Fig.~\ref{fig:size}, we show the AP improvement obtained after correcting each type of localization error. We find that for body and face keypoint localization, the performance can be significantly improved by correcting small-scale human poses, especially the Missing error. For hand pose estimation, errors impact performance more on larger instances. For larger-scale instance, instead of only estimating the rough position, more accurate keypoint localization is required. However, due to the frequent motion blur and severe occlusion (interaction with objects), it is still very challenging to estimating the hand poses of large instances.

	\subsubsection{Qualitative Analysis}
	
\begin{figure*}[tb]
		\centering
		\includegraphics[width=0.9\textwidth]{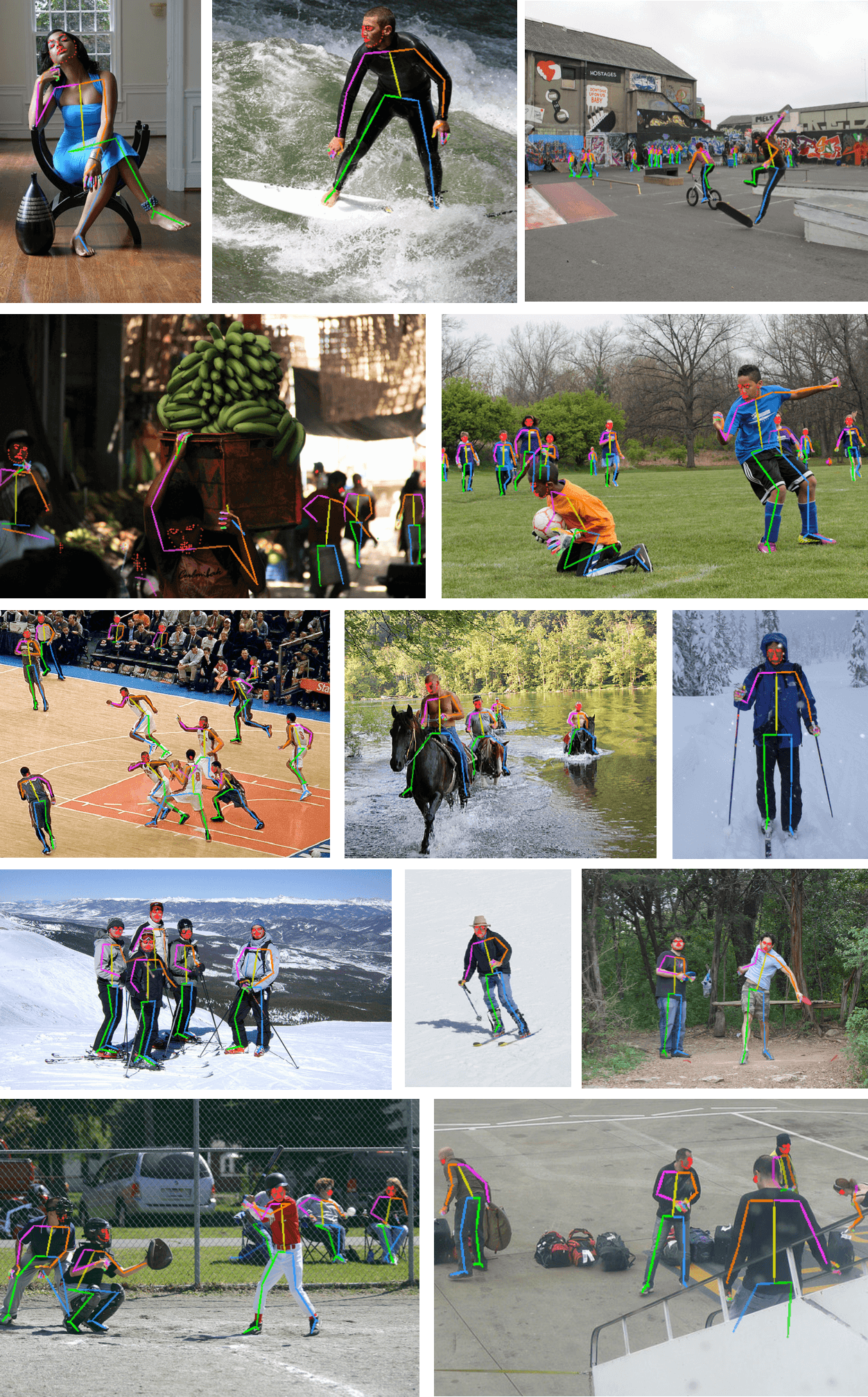}
		\caption{Qualitative evaluation results of our approach in handling challenges including occlusion, close proximity, and small scale persons. 
		}
		\label{fig:qualitative_result}
	\end{figure*}

\begin{figure*}[tb]
		\centering
		\includegraphics[width=0.8\textwidth]{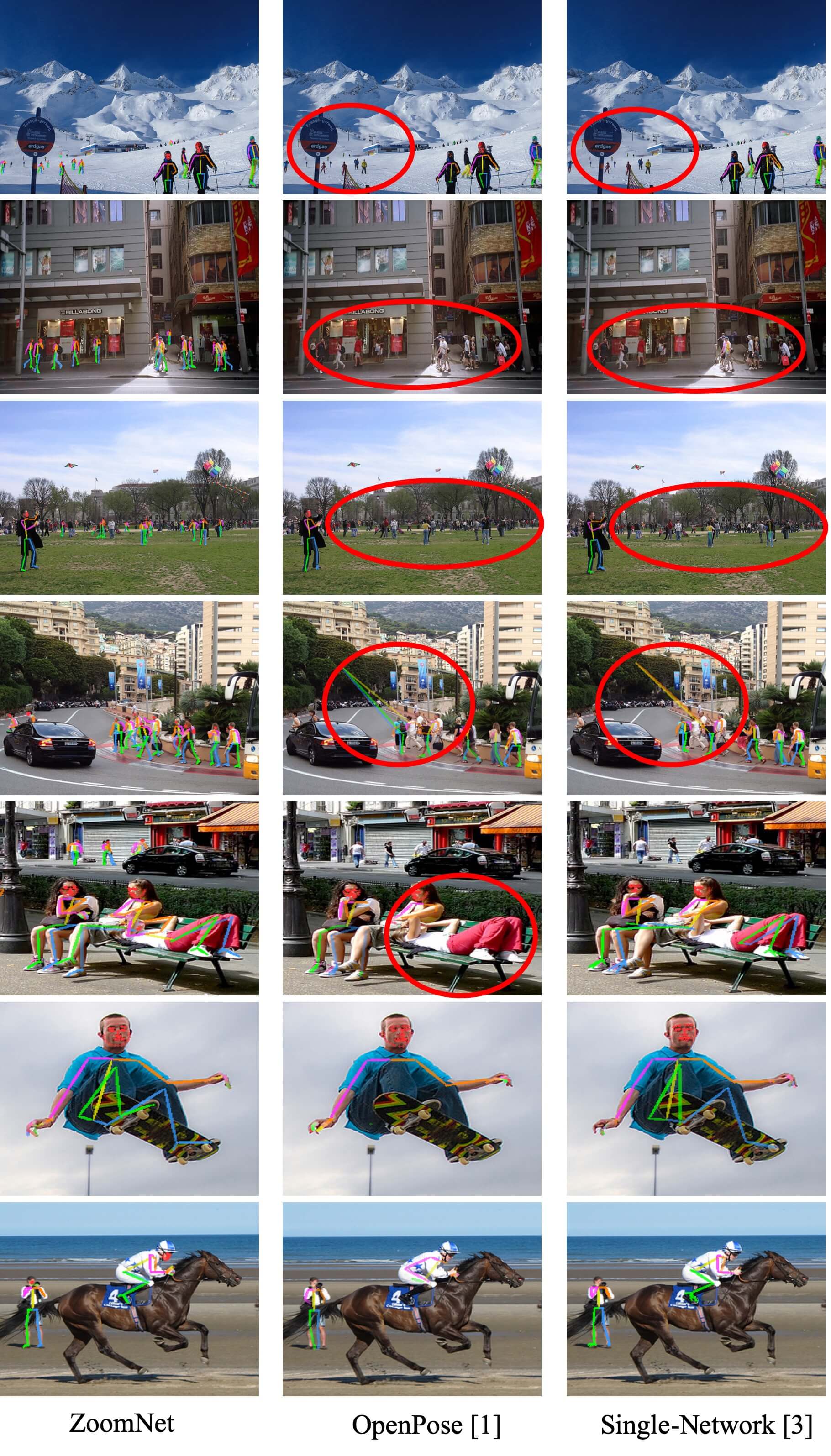}
		\caption{Qualitative comparison between our proposed ZoomNet, OpenPose~\cite{cao2018openpose} and Single-Network~\cite{hidalgo2019single}. Our approach outperforms the state-of-the-art approaches especially on face/hand keypoints and are more robust to scale variance. 
		}
		\label{fig:qualitative_comparison}
	\end{figure*}

	Fig.~\ref{fig:qualitative_result} shows the qualitative evaluation results of our approach, and Fig.~\ref{fig:qualitative_comparison} qualitatively compares the results of ZoomNet, OpenPose~\cite{cao2018openpose} and Single-Network~\cite{hidalgo2019single}. Both of them show the capacity of our proposed ZoomNet in handling challenges including occlusion, close proximity, and small scale persons. We find that our ZoomNet significantly outperforms the previous state-of-the-art method~\cite{hidalgo2019single}, especially for face/hand keypoints. First, we observe that compared to these bottom-up approaches, ZoomNet better handles the small scale problem of human instances (see Line\#1,2,3). Second, we find that the grouping of OpenPose~\cite{cao2018openpose} and Single-Network~\cite{hidalgo2019single} is sometimes erroneous due to lack of human body constraints (see Line\#4). Third, ZoomNet is generally better at localizing the hand/face keypoints with occlusion, pose variations, and small scales (see Line\#6,7). ZoomNet improves upon the state-of-the-art methods by zooming in to the hand area for higher resolution. However, we also find some failure cases of our proposed ZoomNet. We observe that it still has difficulty in dealing with small face/hands with low-resolution and motion blur. 
	
\end{document}